\documentclass[11pt]{article}

\usepackage[T1]{fontenc}
\usepackage[utf8]{inputenc}
\usepackage{libertinus}
\usepackage{amsmath}
\usepackage{libertinust1math}
\usepackage{bm}
\usepackage[varqu,varl,scaled=0.94]{zi4}
\usepackage[a4paper,margin=2.5cm,headheight=14pt]{geometry}
\usepackage{microtype}
\usepackage{graphicx}
\usepackage{booktabs}
\usepackage{tabularx}
\usepackage{array}
\usepackage{xcolor}
\usepackage[font=small,labelfont=bf,skip=5pt]{caption}
\usepackage{enumitem}
\usepackage{listings}
\usepackage[authoryear,round]{natbib}
\usepackage{fancyhdr}
\usepackage{hyperref}
\usepackage{orcidlink}
\usepackage[capitalise,noabbrev]{cleveref}
\usepackage{placeins}

\definecolor{accent}{RGB}{31,78,121}
\definecolor{todofg}{RGB}{160,95,0}
\definecolor{todobg}{RGB}{255,247,230}
\definecolor{codebg}{RGB}{246,246,246}

\hypersetup{
  colorlinks=true, linkcolor=accent, citecolor=accent, urlcolor=accent,
  pdftitle={Bongard: Training Machine Intuition},
  pdfauthor={Li Ding, Haidi Jin, Chen Ji},
  pdfsubject={Technical report},
  pdfkeywords={System One models, intuition, encoder-decoder, T5Gemma 2, JEPA, calibration, reinforcement learning}
}
\setlist{itemsep=2pt,topsep=4pt,parsep=0pt}
\newcolumntype{L}{>{\raggedright\arraybackslash}X}
\newcolumntype{P}[1]{>{\raggedright\arraybackslash}p{#1}}

\newcommand{\sg}{\operatorname{sg}}
\newcommand{\CE}{\operatorname{CE}}
\newcommand{\softplus}{\operatorname{softplus}}
\newcommand{\norm}[1]{\lVert #1\rVert}

\begin{document}

\thispagestyle{plain}
\begin{center}
  {\LARGE\bfseries Bongard: Training Machine Intuition\par}
  \vspace{6pt}
  {\large An Open Encoder--Decoder Model for Probabilistic Judgment\par}
  \vspace{14pt}
  {\large Li Ding\,\orcidlink{0009-0001-6883-0911}\,\textsuperscript{*}
   \qquad Haidi Jin \qquad Chen Ji\par}
  \vspace{6pt}
  {AgentBull Pte Ltd\par}
  \vspace{3pt}
  {\small Technical report \textperiodcentered{} September 2026\par}
\end{center}
{\renewcommand{\thefootnote}{\fnsymbol{footnote}}%
 \footnotetext[1]{Corresponding author: Li Ding, \href{mailto:boris@agentbull.com}{\texttt{boris@agentbull.com}}.}}

\begin{abstract}
\noindent
Human intelligence relies heavily on learned intuition: recognising patterns and judging situations without explicitly unfolding every intermediate step. We introduce \textbf{Bongard}, an open-weight System One model that treats machine intuition as an independent capability to design and train. A T5Gemma~2 4B-4B encoder--decoder separates reading the evidence from making judgments. The encoder reads the state bidirectionally together with the question instructions, and separate decoder branches share this encoding, so many judgments about the same situation require only one reading of the state. A trained head returns probabilities over the supplied candidates without generating text. Training proceeds in three stages, from supervised judgments to semantic relationships to action outcomes, and each stage updates all 7.09 billion trainable parameters on one Blackwell GPU. Joint-embedding post-training raises accuracy on held-out rephrasings from 75.7\% to 85.9\%. A sandbox stage then learns outcome distributions from action rollouts and exact oracles, raising accuracy on a frozen sandbox panel from 50.6\% to 64.8\%. On DecisionBench, the final model reaches 78.05\% accuracy over 23,900 decisions and ranks fourth of 61 systems in the public comparison. On one RTX PRO 6000, its median latency is 36~ms for short requests, and 32 questions about one state take 221~ms. Bongard demonstrates that machine intuition can be systematically trained via representation learning and outcome feedback, providing an open, efficient alternative for high-throughput decision workloads.

\medskip
\noindent\textbf{Weights:} \href{https://huggingface.co/AgentBull/bongard-mini}{\texttt{huggingface.co/AgentBull/bongard-mini}}
\end{abstract}

\section{Introduction}\label{sec:intro}

Human expertise often manifests as intuitive judgment: rapidly identifying patterns and evaluating situations without explicitly verbalising intermediate reasoning steps \citep{kahneman2011thinking}. Expert intuition develops when experience exposes useful regularities and provides feedback \citep{kahneman2009conditions}. Its value extends beyond speed, because an intuitive judgment can draw on a whole pattern whose relevant features are difficult to state one by one. Explicit analysis, by contrast, can change which features a person attends to: in preference studies, asking people to analyse their reasons reduced agreement with expert judgments \citep{wilson1991thinking}.

In machine learning, related phenomena emerge when models internalise structured tasks. Transformers trained on chess positions can play at grandmaster level without search \citep{ruoss2024chess}, and models trained on Othello move sequences can learn internal board representations \citep{li2023othello}. In both cases, training places structure in the model, which then uses it directly at inference time.

We study \emph{machine intuition} as the ability to learn relationships within situations and use them to form direct judgments. Bongard exposes this capability through a programmable interface. A request supplies evidence, questions and their possible outcomes, and the model returns a probability distribution for each question without generating an intermediate explanation. The interface covers judgments of meaning and relevance as well as the consequences of actions, so the same model can assess a document, detect conflicting facts or judge an agent's next step.

TypeSafe AI named models for this workload \emph{System One models}, after Kahneman's System~1, and released Jev as the first of them \citep{typesafe2026systemone}. A System One model takes a state and typed questions and returns a separate distribution over the outcomes of each question. Although the category is recent, it already serves production traffic, for example row by row within SQL queries \citep{motherduck2026} and as the step executor of browser and mobile agents \citep{jevultrafast2026,zhang2026jevmobile}. Thousands of public projects use such models for attribute judgment, scoring, action selection, content filtering and model or tool selection \citep{ling2026jevwild}. Replacing large language model (LLM) calls with them can reduce both latency and cost: in one edge service, Jev cut median decision latency by 15.9--26.5\% and API fees per correct completion by 69.0--70.6\% \citep{li2026edgejev}.

The same interface can expose very different models. Some open implementations wrap pretrained models and read probabilities from answer tokens \citep{openjev2026,semif2026}. Others train a decision model: Kev adds LoRA adapters and a candidate-scoring head to Qwen \citep{palmer2026kev}, whereas JevK5 and Plumb-4B train adapters but retain letter-logit readouts \citep{jevk52026,plumb2026}. These choices determine what is trained, how the evidence is represented and how candidates are scored, and they imply different costs of adaptation.

\begin{figure}[t]
  \centering
  \includegraphics[width=\linewidth]{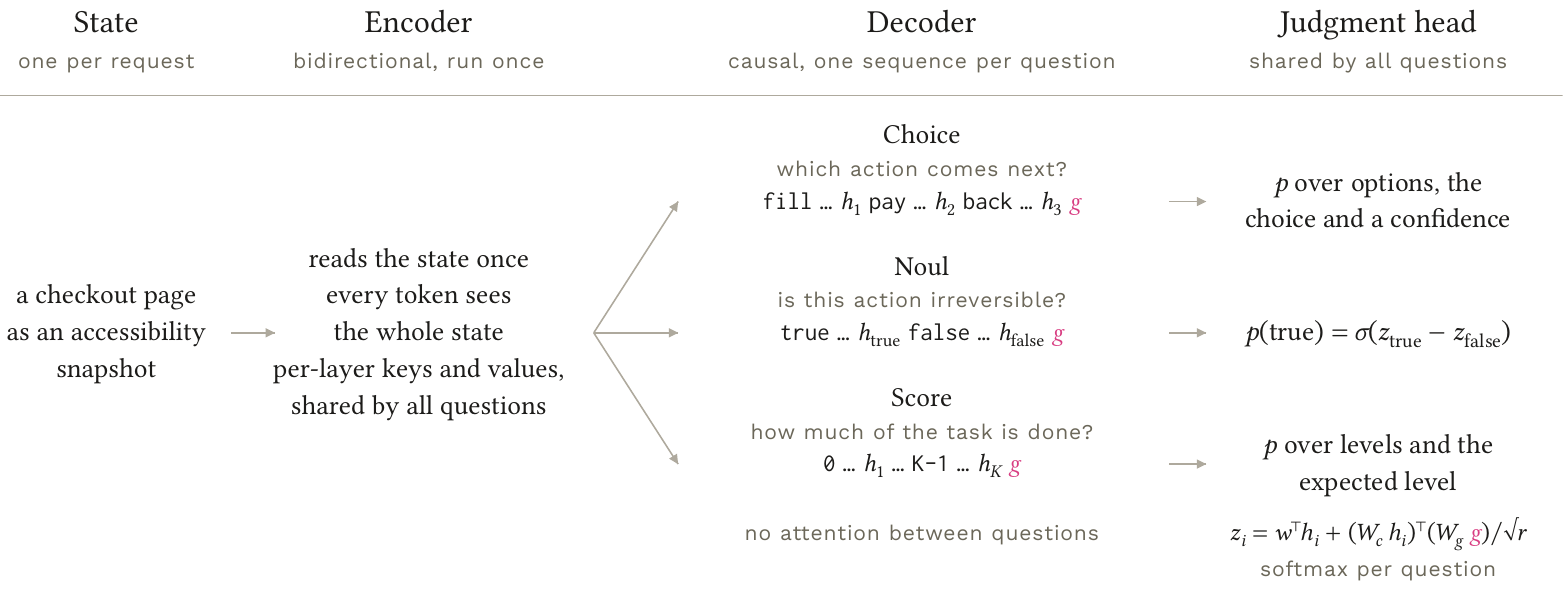}
  \caption{Architecture of Bongard: several typed questions share a single encoding of the state, and no tokens are generated. In this example, three questions of different types concern one checkout page. The encoder reads the state once, and each question is a separate decoder sequence. The decoder processes all sequences in one batched pass, with merged self- and cross-attention over the shared state keys and values. There is no attention between questions. Hidden states at the candidate-end markers ($h_i$) and the decision-end marker ($g$) feed a shared bilinear judgment head.}
  \label{fig:arch}
\end{figure}

Bongard combines a T5 encoder--decoder judgment architecture with training on semantic relationships and action outcomes (\cref{fig:arch}). The encoder forms a bidirectional representation of the evidence with the question instructions in view. Separate decoder branches judge the supplied candidates from this shared representation, and a dedicated head returns their probabilities directly. This structure suits situations that require several judgments from related evidence, such as a contract, an incident log or a page snapshot. Machine intuition thus becomes a capability that can be trained across tasks and called from within an application.

T5Gemma~2 supplies the encoder--decoder backbone \citep{zhang2025t5gemma2}. It inherits Gemma~3 pretraining \citep{gemma3}, supports a 128K-token context and more than 140 languages, and includes a vision encoder. Bongard adds a judgment head of 1.3 million parameters that reads its hidden states. The model is trained in three stages, each on one GPU (\cref{fig:pipeline}). After supervised training on judgments, a joint-embedding predictive architecture (JEPA) objective aligns the representations of semantically related inputs, and a final sandbox stage learns from the outcomes of actions in executable environments.

Our contributions are the following.
\begin{itemize}
  \item \textbf{An encoder--decoder architecture for machine intuition} (\cref{sec:task,sec:arch}). A bidirectional encoder represents the shared evidence once, and separate decoder branches make probabilistic judgments from it, so $N$ questions cost one encoder pass and about $\lceil N/8 \rceil$ decoder calls. This T5 structure distinguishes Bongard from the decoder-only and diffusion systems in \cref{tab:compare}. Probes with a retention rule fixed in advance favoured the unmodified backbone over every change we tried.
  \item \textbf{Representation alignment and outcome learning} (\cref{sec:jepa,sec:rl}). Stage~2 exposes a gap between accurate decisions and the representation behind them, and it shows why cosine alignment saturates without a next-token anchor. A pack-centred contrastive objective makes weak content correspondences recoverable from the decision representation, and the complete stage improves consistency across re-expressions of the same questions. Stage~3 grounds judgments in environment transitions: rollouts and exact oracles label every candidate action with an outcome distribution, and a direct proper-scoring loss trains the model on these distributions.
  \item \textbf{Full-model post-training without a GPU cluster} (\cref{sec:infra}). We post-train all 7.09 billion trainable parameters on one Blackwell GPU, with NVFP4 and FP8 matrix multiplications, FlashAttention-4 over unpadded packs and 8-bit AdamW. The three stages took about 51, 23 and 11 hours, so our training pipeline requires no cluster-scale infrastructure.
\end{itemize}

The final model is open at \href{https://huggingface.co/AgentBull/bongard-mini}{\texttt{huggingface.co/AgentBull/bongard-mini}} under the Gemma terms of use. The runtime and the server are open at \href{https://github.com/AgentBull/bongard}{\texttt{github.com/AgentBull/bongard}} under the Apache~2.0 licence. The name refers to Bongard problems \citep{bongard1970pattern}, visual puzzles in which a hidden rule separates two sets of figures. Solving them requires recognising invariant discriminative rules directly from examples---a core objective of System One models.

\section{Task and interface}\label{sec:task}

A request contains a \emph{state}, optional images and a dictionary of typed \emph{questions} (\cref{tab:primitives}). The state is a string, a JSON object or an array. The \emph{candidates} of a question are its possible outcomes: Choice options, the two Noul outcomes or Score levels. Questions may ask for a label, such as the queue for a ticket, or for an outcome, such as whether an action is irreversible. A request may mix all three question types.

Requests and responses follow TypeSafe's System One API \citep{typesafe2026docs}, so existing clients work unchanged. Images are a Bongard extension. For each question, the response gives the distribution over its candidates and derived fields (\cref{app:interface}). These fields are the selected option and a confidence for Choice, the truth probability for Noul, and the expected level and a confidence for Score.

\begin{table}[t]
  \centering
  \caption{The three question primitives. The same head answers all of them in the same forward pass.}
  \label{tab:primitives}
  \small
  \begin{tabularx}{\linewidth}{@{}lLLL@{}}
    \toprule
    Primitive & Question & Criteria & Answer \\
    \midrule
    \textbf{Choice} & Which named alternative applies? & 1--255 options, each with an optional description & A probability per option, the selected option and a confidence \\
    \textbf{Noul} & Is this statement true? & Optional descriptions of the true and false outcomes & The probability that the statement is true \\
    \textbf{Score} & Where does the input fall on an ordered scale? & 2--10 ordered level descriptions & A probability per level, the expected level and a confidence \\
    \bottomrule
  \end{tabularx}
\end{table}

Two properties of this interface shape the design. First, the request fixes the outcome space of each question, so an answer cannot leave its type. Second, questions are isolated: for fixed weights $\theta$,
\begin{equation}
  \text{answer}_i = F_\theta(\text{state}, I, \text{question}_i), \label{eq:isolation}
\end{equation}
where $I$ is the list of distinct question instructions that the encoder reads with the state (\cref{sec:parallel}). Other questions can reach an answer only through $I$, never through their names, candidates or criteria. For the stage-1 and stage-2 checkpoints, $I$ is empty. Applications can therefore send every question they have about a state in one request.

\section{Architecture}\label{sec:arch}

Bongard reads a state once and answers many questions about it (\cref{fig:arch}). It has three parts: a bidirectional encoder, a causal decoder with one sequence per question, and a compact judgment head.

\subsection{Architecture follows the decision workload}\label{sec:why}

Bongard uses T5Gemma~2 as a judgment model. The encoder represents the situation once, and for each question a decoder branch reads this representation before the judgment head scores the candidates directly. This division separates learning to read the evidence from learning to apply it in a judgment.

\textbf{Adaptation from decoder-only pretraining.} T5Gemma~2 is adapted from the decoder-only Gemma~3 with a UL2-style objective \citep{tay2023ul2,zhang2025t5gemma2}. Both halves therefore inherit the pretraining, the 262K-entry vocabulary and the vision tower of Gemma~3. With the same knowledge, it matches or exceeds Gemma~3 after pretraining, clearly outperforms it after post-training and is stronger on long contexts \citep{zhang2025t5gemma2}. The first T5Gemma models showed a better quality--efficiency trade-off than their decoder-only originals \citep{zhang2025encdecgemma}. RedLLM finds that instruction-tuned encoder--decoders match or exceed decoder-only models up to 8B parameters, with substantially more efficient inference \citep{zhang2025redllm}.

\textbf{Bidirectional evidence encoding.} An encoder represents every state token in the context of the whole state, whereas a decoder-only model reads its prefix forward, so early tokens never attend to what follows them. Whole-state context suits tables, contracts, logs and page snapshots, whose cells and lines depend on headers and neighbours. In the controlled comparison of \citet{raffel2020t5}, an encoder--decoder with a denoising objective transferred better than decoder-only and prefix language models of similar cost. \citet{su2023decoderonly} argues for the opposite design, on the grounds that bidirectional attention matrices tend towards low rank \citep[see also][]{dong2021rank}. In our probe, however, forcing the frozen T5Gemma~2 1B-1B encoder to read forward only cut accuracy from 69.8\% to 64.6\% and raised NLL more than fourfold (\cref{tab:probes}a).

\textbf{Single-pass state encoding with shared cross-attention.} The encoder reads the state, typically hundreds to tens of thousands of tokens, only once. Each decoder layer then projects the encoded state into keys and values once and shares them with every question, as in Fusion-in-Decoder \citep{izacard2021fid}. A decoder-only model can recover this reuse only at serving time, by prefilling the state as a shared prefix and batching the per-question suffixes \citep{juravsky2024hydragen,semif2026,jevfire2026,openjevsglang2026}, and its shared representation remains causal. The split into two stacks costs memory rather than compute: each token passes through only one 34-layer stack shaped like Gemma~3 4B.

\textbf{Asymmetric capacity allocation.} In a decoder-only model, every question token passes through the full stack. In an encoder--decoder, the state passes once through the encoder, and question tokens pass only through the decoder. Because adaptation allows the two stacks to differ in size \citep{zhang2025encdecgemma}, capacity can shift to the encoder, which runs once per state, while the decoder, which runs for every question, remains small. A decoder-only model offers no such trade-off, since every token incurs the cost of the full stack. The current model uses the balanced 4B-4B configuration and leaves unbalanced configurations to future work.

\textbf{Trained judgment head instead of vocabulary readout.} Next-token logits over letters, yes/no tokens or masked slots give a distribution over vocabulary items. This distribution is entangled with the tokeniser, label-word priors and option position \citep{zhao2021calibrate,zheng2024mcq}, which post-hoc temperature scaling addresses only in part \citep{guo2017calibration}. Readouts over fixed option codes avoid the vocabulary but still score code slots rather than the candidates themselves \citep{garg2026imajev}. Bongard instead scores candidates given in full, by name and description, with a head trained under proper scoring rules. Encoder--decoder rerankers established the pattern of scoring from decoder outputs \citep{nogueira2020monot5,zhuang2023rankt5}, and Bongard extends it to typed judgments with many questions per request.

\subsection{Backbone and judgment head}\label{sec:backbone}

Bongard keeps the architecture of \texttt{google/t5gemma-2-4b-4b} unchanged (\cref{tab:config}) and adds new parameters only for the head and the multimodal projector. The state is serialised as compact JSON and encoded together with any image tokens. Each question is compiled into a separate decoder sequence that spells out its type, instructions and candidates (\cref{app:interface}). Two unused entries of the existing vocabulary serve as markers: a candidate-end marker follows each candidate, and a decision-end marker ends the question. The vocabulary is therefore never resized. A guarded tokeniser encodes any literal marker text in user input as byte tokens, so user input cannot forge a marker.

\begin{table}[t]
  \centering
  \caption{Bongard 4B-4B configuration. Training updates all components except the vision tower, so 7.09B of the 7.51B parameters are trainable.}
  \label{tab:config}
  \small
  \begin{tabularx}{\linewidth}{@{}lLr@{}}
    \toprule
    Component & Configuration & Parameters \\
    \midrule
    Text encoder & 34 layers, $d = 2560$, FFN 10,240, 8 query / 4 KV heads of dimension 256, window 1,024 (5 local : 1 global), bidirectional & 3.21B \\
    Text decoder & Same shape, causal, merged self- and cross-attention & 3.21B \\
    Tied embeddings & 262,144 tokens $\times$ 2,560 & 0.67B \\
    Vision tower & SigLIP \citep{zhai2023siglip}, 896 px, 256 tokens per image, frozen & 0.42B \\
    Projector, head & Trained, with $w \in \mathbb{R}^{2560}$ and $W_c, W_g \in \mathbb{R}^{256 \times 2560}$ & 4.26M \\
    \midrule
    \textbf{Total} & Bongard limit: 32,768 tokens per state plus question & \textbf{7.51B} \\
    \bottomrule
  \end{tabularx}
\end{table}

Let $h_i$ be the last-layer decoder hidden state at the $i$-th candidate-end marker, and let $g$ be the hidden state at the decision-end marker. We call $h_i$ a \emph{candidate readout} and $g$ the \emph{decision readout}. The head scores candidate $i$ as
\begin{equation}
  z_i = w^\top h_i + \frac{(W_c h_i)^\top (W_g\, g)}{\sqrt{r}}, \qquad r = 256, \label{eq:head}
\end{equation}
and a softmax over the scores $z_1, \dots, z_K$ of the $K$ candidates gives the distribution of the question.

For fixed $g$, the head is a linear scorer whose weights depend on the whole question. This dependence lets a causal decoder score early candidates with information from later candidates. The interaction must be multiplicative, because an additive term in $g$ would cancel in the softmax. Noul returns $\sigma(z_{\text{true}} - z_{\text{false}})$ and Score returns the level distribution and its expectation, so no primitive requires token generation.

\subsection{Parallel questions and question-aware encoding}\label{sec:parallel}

T5Gemma~2 normalises decoder self-attention and cross-attention in one softmax. Bongard computes this softmax in two partitions: the causal prefix of each question and the shared state. Following Hydragen \citep{juravsky2024hydragen}, it merges the two partitions with their log-sum-exp weights, $\log Z_q$ for the question prefix and $\log Z_s$ for the state. The state partition batches the queries of all questions against a single copy of the state keys and values. For a given encoder input, a batched request therefore computes exactly the same function as separate single-question requests. The decoder processes questions in length-sorted groups of up to eight, so $N$ questions cost one encoder pass and about $\lceil N/8 \rceil$ decoder calls.

\textbf{Question-aware encoding.} From stage~3 on, the encoder reads the state followed by the distinct text instructions of the request's questions. The encoder can therefore focus its reading on what the questions ask, while candidates and criteria stay in the decoder. A request that would exceed the token budget with the instructions is encoded without them. This rule is deterministic, so training and serving agree.

Tests confirm isolation directly. Without question-aware encoding, answers stay the same when other questions in the request are reordered, renamed, deleted or injected. With it, the same holds for every change that keeps $I$, because the encoder input is then unchanged. Gradients from one question also never reach the decoder sequence of another question.

An optional cache of encoder outputs serves repeated requests about a recent state. A cache hit needs the same encoder input, so under question-aware encoding it also needs the same $I$. On a 1B-1B model, this cache cut the latency of a repeated request with a 1,924-token state tenfold, from 1,165 to 115~ms, with bit-identical logits.

\subsection{Architecture probes}\label{sec:validation}

We probed the design on the 1B-1B model of the same family (\cref{tab:probes}). The native bidirectional encoder outperformed the forward-only mask and the mask with separate forward and backward heads on every measure. Its final hidden states also kept the highest effective rank. We therefore found no sign of the rank collapse that \citet{su2023decoderonly} predicts.

We retained none of the four decoder modifications. Bidirectional global layers within a question improved held-out negative log-likelihood (NLL) on both seeds, but by widely varying amounts. They also worsened transfer on one seed and so failed the retention rule that we fixed before training. A tail readout added 13--25\% to latency and transferred worse. The native fusion of self- and cross-attention is already the gate $\sigma(\log Z_s - \log Z_q)$, and a learned bias on this gate hurt held-out NLL. The value residual of ResFormer \citep{zhou2024valueresidual}, which RWKV-7 also uses \citep{peng2025rwkv7}, raised validation NLL on both seeds.

\begin{table}[t]
  \centering
  \caption{Architecture probes on T5Gemma~2 1B-1B. (a) Encoder masks with a frozen backbone and a trained head, on synthetic judgments of thresholds, negation and role binding, averaged over two seeds. Fact direction is the share of fact changes that move the truth probability in the correct direction. (b) Decoder modifications after short full-parameter training, as changes relative to the unmodified model for seeds 23 and 37. The value-residual row comes from a longer run and reports cross-task NLL as transfer.}
  \label{tab:probes}
  \small
  \begin{tabularx}{\linewidth}{@{}Lrrrrr@{}}
    \toprule
    \multicolumn{6}{@{}l}{\textit{(a) Encoder attention mask}} \\
    & Accuracy $\uparrow$ & NLL $\downarrow$ & Brier $\downarrow$ & Fact direction $\uparrow$ & Effective rank \\
    \midrule
    Bidirectional (native) & \textbf{69.8\%} & \textbf{0.677} & \textbf{0.219} & \textbf{100.0\%} & \textbf{49.7} \\
    Forward only & 64.6\% & 2.944 & 0.320 & 70.8\% & 37.9 \\
    Forward and backward heads & 57.3\% & 1.596 & 0.363 & 62.5\% & 31.7 \\
    \midrule
    \multicolumn{6}{@{}l}{\textit{(b) Decoder modification}} \\
    & \multicolumn{2}{r}{Held-out NLL $\Delta$ $\downarrow$} & \multicolumn{2}{r}{Transfer NLL $\Delta$ $\downarrow$} & Latency $\Delta$ \\
    \midrule
    Bidirectional global layers & \multicolumn{2}{r}{$-25.4\%$ / $-1.0\%$} & \multicolumn{2}{r}{$-3.3\%$ / $+1.7\%$} & $\approx 0\%$ \\
    Tail readout, zero-initialised residual & \multicolumn{2}{r}{$-4.3\%$ / $+0.2\%$} & \multicolumn{2}{r}{$+10.7\%$ / $+12.1\%$} & $+13$ to $+25\%$ \\
    Zero-initialised fusion bias & \multicolumn{2}{r}{$+26.4\%$ / $+27.5\%$} & \multicolumn{2}{r}{$-6.9\%$ / $+45.2\%$} & $\approx +1\%$ \\
    Value residual & \multicolumn{2}{r}{$+3.2\%$ / $+9.3\%$} & \multicolumn{2}{r}{$-0.03\%$ / $-1.2\%$} & -- \\
    \bottomrule
  \end{tabularx}
\end{table}

\subsection{Jev, open implementations and design tradeoffs}\label{sec:compare}

\Cref{tab:compare} locates Bongard within the System One ecosystem. An inference wrapper changes how an existing model is called and how its answer is read, whereas training changes the judgment function itself. Bongard trains the text backbone, the multimodal projector and the judgment head on judgments, semantic correspondences and action outcomes, while the vision tower stays frozen.

Kev makes the architectural contrast concrete \citep{palmer2026kev}. It trains rank-16 LoRA adapters and a pointer head on Qwen, and its server reuses a causal state cache across question branches. Bongard instead forms a bidirectional state representation in a T5 encoder, which its decoder branches share. Full-model updates with additional representation and outcome objectives require more training and data construction than adapter training, but they allow Bongard to shape the evidence representation and the judgment function together.

In practice, bidirectional encoding allows earlier facts to be represented in the context of later ones, which matters for relationships across tables, documents and page snapshots. Separate encoder and decoder stacks also allow future designs to allocate more capacity to the single reading of a state than to each repeated question. Semantic-correspondence training targets invariance to changes in wording and modality (\cref{sec:jepa}), whereas outcome training ties judgments to the effects of actions (\cref{sec:rl}).

Jev is the closed reference system. TypeSafe describes a specialised architecture and a training method, Reinforcement Learning for Calibrated Decisions (RLCD) \citep{typesafe2026systemone}, and API experiments suggest shared state processing, isolated question branches and interaction between candidates \citep{hume2026jev}. Its internal representation and full training recipe, however, are not public. Bongard offers an open encoder--decoder route to the same class of callable judgments, and the task-level comparisons in \cref{tab:bench,fig:same-items} show where the strengths of the two systems differ.

\begin{table}[t]
  \centering
  \caption{Architecture and learning choices in documented System One systems. Sources: Jev \citep{typesafe2026systemone,hume2026jev}, OpenJev \citep{openjev2026}, Kev \citep{palmer2026kev}, JevK5 and Plumb-4B \citep{jevk52026,plumb2026}, AutoJev \citep{autojev2026}, MoJev \citep{mojev2026}, CLM \citep{clm2026}, imajev \citep{garg2026imajev}, and Laya and Verdict \citep{laya2026,openjev2026}. Jev's architectural details are inferred from API behaviour.}
  \label{tab:compare}
  \small
  \begin{tabularx}{\linewidth}{@{}P{2.2cm}LLL@{}}
    \toprule
    System & Backbone and training & How the state is read & How the answer is read \\
    \midrule
    \multicolumn{4}{@{}l}{\textit{Encoder--decoder}} \\
    \textbf{Bongard} & T5Gemma~2 4B-4B; full text-model training on judgments, semantic pairs and action outcomes & Bidirectional, question-aware encoder, once per request, shared by separate decoder branches & Trained bilinear head on candidate and decision readouts, no sampling \\
    \midrule
    \multicolumn{4}{@{}l}{\textit{Closed}} \\
    Jev & Undisclosed backbone; RLCD described by TypeSafe & Isolated question branches (observed), state processed once (suggested by latency) & Direct probabilities, options interact (observed) \\
    \midrule
    \multicolumn{4}{@{}l}{\textit{Decoder-only or diffusion language models}} \\
    OpenJev default & Pretrained DiffusionGemma 26B-A4B; inference wrapper & State and all questions in one canvas & Label-token probabilities at masked slots, noisy re-reads when uncertain \\
    Kev & Qwen3.5/3.8; rank-16 LoRA and pointer head, supervised decision loss & Causal state cache reused across separate question rows in the server & Trained pointer head scores each option against the final decision state \\
    JevK5, Plumb-4B & Qwen3.5-4B decoder with LoRA adapters & Causal prefix, one prompt per question & Letter logits over options with a fitted temperature \\
    AutoJev-27B & Qwen3.8-27B decoder, full fine-tuning & Causal prefix, one pass per question & Probabilities over supplied choices, scalar temperature \\
    MoJev & Qwen3.5-0.8B decoder & Tree-packed attention over state and questions & Shared rank-512 judgment head \\
    CLM & Frozen Qwen3-8B with small projection heads & State embedded once, as one vector & Softmax over scaled cosine of state and action projections \\
    imajev & Qwen3.5 2B, 4B or 9B decoder with LoRA adapters (rank 16) & Causal prefill once per request (up to 4,096 tokens), one pass per question & Readout over 255 option codes plus an unknown option, four option rotations when served \\
    \midrule
    \multicolumn{4}{@{}l}{\textit{Text encoders}} \\
    Laya, Verdict & ModernBERT encoders & Bidirectional, short inputs, one pass per question & Classification or GLiClass head \\
    \bottomrule
  \end{tabularx}
\end{table}

\section{Training}\label{sec:training}

\begin{figure}[t]
  \centering
  \includegraphics[width=\linewidth]{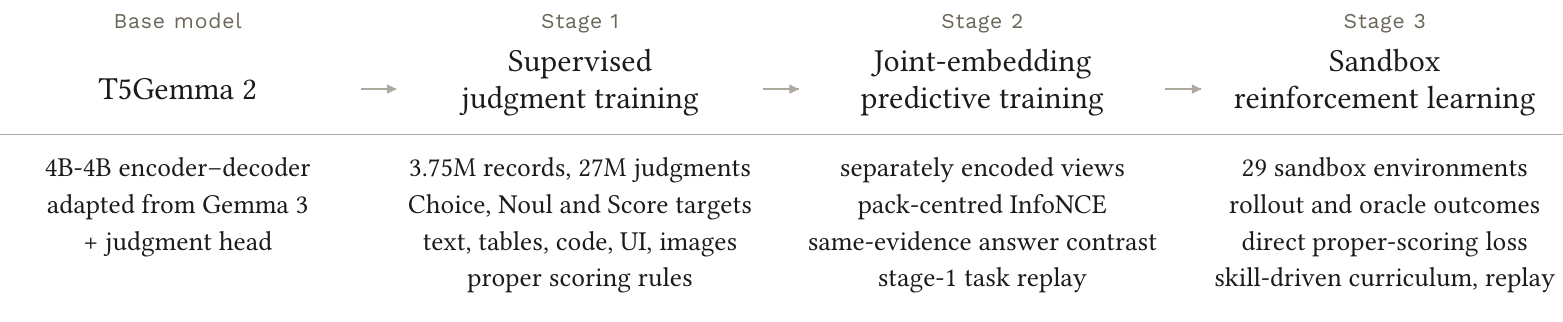}
  \caption{Training pipeline. A pretrained T5Gemma~2 encoder--decoder receives a judgment head and is then trained in three stages: supervised judgment training, joint-embedding predictive training and sandbox reinforcement learning. Each stage updates all trainable parameters on one GPU.}
  \label{fig:pipeline}
\end{figure}

The three training stages (\cref{fig:pipeline}) progressively establish task competence (stage~1: supervised training), semantic alignment (stage~2: JEPA) and outcome-driven calibration (stage~3: sandbox RL). All stages use the same request format and judgment head, and each trains on one GPU (\cref{sec:infra}). From stage~3 on, the encoder also reads the question instructions (\cref{sec:parallel}).

\subsection{Stage 1 (supervised training): learning to judge}\label{sec:sft}

Stage~1 draws supervised judgments from text, structured data and images (task families in \cref{app:data}). Each record is a state with one or more questions and a target for each question. Noul targets are hard or soft Bernoulli targets, and Choice and Score targets are hard classes, full distributions or \emph{allowed sets} of tied correct candidates. All losses apply proper scoring rules to the output distributions of the head \citep{gneiting2007strictly}. Four construction methods matter most.

\textbf{Program-exact synthesis.} Generators create constrained worlds and executable problems, and exact solvers or executions label the questions. Where independent checks are available, a label is kept only if they agree. The generators also create paired examples that either change the answer through a relevant fact or preserve it under an irrelevant change.

\textbf{Derived judgments.} A labelled example can support related judgments, including per-option truth, numeric thresholds, joint events and next-step prediction. Each question sees only the evidence available before its target outcome.

\textbf{Retrieval judgments.} Retrieval examples become listwise, pairwise and single-passage questions. A teacher reranker filters ambiguous examples, while the supervised targets remain hard. Candidate order varies, and some questions have no relevant passage. The retrieval intent stays in the state because relevance depends on it.

\textbf{Structured data and teachers.} Structured records support lookup and prediction questions. Rule-based checks validate exact answers, and consistency checks filter teacher labels. Where exact labels are unavailable, filtered teacher distributions provide soft targets. Questions with known random mechanisms carry exact probabilities.

Across the corpus, instructions are rewritten in multiple languages and candidate order is varied. Related records stay in the same split because split groups are assigned before rewriting or pairing. Stage~1 makes one pass over the corpus.

\subsection{Stage 2 (JEPA): learning the structure behind judgments}\label{sec:jepa}

\subsubsection{Motivation}\label{sec:jepa-why}

Supervised training constrains only the final output distribution, leaving the model prone to relying on template fingerprints, option heuristics or surface shortcuts. To ensure robust generalisation, the underlying representation should reflect semantic invariants: equivalent facts phrased differently should map to similar embeddings, while the prediction of an outcome should align with the representation of the observed outcome.

JEPAs learn such invariants: they predict the representation of one view from another view in embedding space \citep{lecun2022path,dawid2023lvebm,assran2023ijepa,bardes2024vjepa,assran2025vjepa2}. LLM-JEPA adds an embedding-space term to the standard loss of a language model \citep{huang2025llmjepa}, and BERT-JEPA applies the idea to encoder sentence embeddings \citep{gillin2026bertjepa}. In LLM-JEPA, however, the hidden state must also predict the next token over a vocabulary of about 260,000 entries. This next-token loss anchors the representation, so a plain cosine term suffices. A judgment model lacks such an anchor: its head reads $g$ only through a 256-dimensional projection, and pure alignment without negatives has nothing to keep representations apart \citep{wang2020alignment}.

\subsubsection{Views}\label{sec:jepa-views}

A JEPA pair consists of two \emph{views}, each a state with a question. The views must agree: a declared family of questions has the same answer distribution under both views. Each pair comes from one stage-1 record, its \emph{origin}, which receives one relation (\cref{tab:relations}). Two kinds of relation carry the alignment terms.

\emph{Content and world pairs} encode their two views separately, from different inputs. For example, a query pairs with a passage's content, and an image pairs with its annotated content. A world and an action pair with the observed next state, and a program pairs with its re-executed result. Two independent solvers or a re-execution verify the targets of these pairs. \emph{Answer states} pair a multiple-choice question with two event questions under the same evidence. One asks whether the correct option is the answer, and the other asks the same of a wrong option.

The alignment terms always act on the decision readout $g$ of each view. As in LLM-JEPA, the decoder itself serves as the predictor, with tied weights. There is no separate predictor, momentum teacher or negative queue.

\subsubsection{Diagnosis: cosine alignment has almost no signal}\label{sec:jepa-geometry}

We probed the stage-1 model with forward passes only, over 256 origins per family, and the results shaped the objective (\cref{tab:geometry}).

\begin{table}[t]
  \centering
  \caption{Geometry of the decision readout $g$ after stage~1, from forward passes over 256 training origins per family. Raw cosine between unrelated readouts is already 0.93--0.996. Centring each side on its mean exposes the true alignment. Logical complements and inherited views are already aligned. Separately encoded content is not, except for paraphrase pairs, whose two sentences share most of their words.}
  \label{tab:geometry}
  \small
  \begin{tabular}{@{}lccr@{}}
    \toprule
    Family & Raw cosine & Centred cosine & Centred retrieval \\
           & paired / unpaired & paired / unpaired & top-1 ($N = 256$) \\
    \midrule
    \multicolumn{4}{@{}l}{\textit{Shared encoded state}} \\
    Answer state & 0.995 / 0.993 & 0.46 / $-0.01$ & 28.9\% \\
    Logical complement & 1.000 / 0.996 & 0.94 / 0.05 & 99.6\% \\
    Inherited view & 1.000 / 0.979 & 0.98 / 0.00 & 71.9\% \\
    \midrule
    \multicolumn{4}{@{}l}{\textit{Separately encoded}} \\
    Paraphrase & 1.000 / 0.996 & 0.86 / 0.02 & 93.0\% \\
    Question $\to$ answer content & 0.936 / 0.929 & 0.30 / 0.05 & 0.8\% \\
    Image $\to$ content & 0.962 / 0.961 & 0.20 / 0.00 & 3.5\% \\
    Query $\to$ passage & 0.950 / 0.947 & 0.23 / $-0.01$ & 11.3\% \\
    World prediction & 0.980 / 0.979 & 0.07 / 0.00 & 1.2\% \\
    Causal prediction & 0.988 / 0.988 & 0.19 / 0.00 & 1.6\% \\
    Assignment (action values) & 0.993 / 0.992 & 0.11 / 0.00 & 1.2\% \\
    Matching (stable allocation) & 0.995 / 0.995 & 0.00 / 0.00 & 0.0\% \\
    \bottomrule
  \end{tabular}
\end{table}

\textbf{Raw cosine is saturated.} All readouts share one dominant direction, the anisotropy that is familiar from contextual representations \citep{ethayarajh2019contextual}. For answer states, $1 - \cos$ is 0.0051 for true pairs and 0.0071 for unrelated pairs. A cosine objective therefore spends nearly all its gradient on making all readouts more collinear. The head's subspace captures only 11.9\% of the item-specific variance of $g$, which is about what a random 256-dimensional subspace would capture.

\textbf{Shared-state views are already aligned, but separately encoded content is not.} After centring, logical complements and inherited views, which share an encoded state, reach retrieval of 72--99.6\%. Further alignment of these views would reward copying the fingerprint of the state. Separately encoded content reaches only 0--11\%, and LLM-JEPA operates in this regime. Paraphrase pairs are the exception: they are encoded separately but reach 93.0\%, because their two sentences share most of their words.

\textbf{The decision readout does not encode the answer.} Although the model answered these multiple-choice questions with 99.6\% accuracy, $g$ was closer to the wrong option's ``no'' event (centred cosine 0.54) than to the correct option's ``yes'' event (0.46). The candidate readouts and the bilinear head decide the answer, and a positive-only objective cannot separate events that share their state and template.

\subsubsection{Objective}\label{sec:jepa-objective}

Stage~2 minimises
\begin{equation}
  \mathcal{L} = \mathcal{L}_{\text{SFT}} + \mathcal{L}_{\text{view}} + \lambda \left( \mathcal{L}_{\text{content}} + \mathcal{L}_{\text{answer}} \right), \label{eq:jepa-total}
\end{equation}
where $\mathcal{L}_{\text{SFT}}$ is the stage-1 judgment loss and $\mathcal{L}_{\text{view}}$ supervises the alternative views. The two alignment terms act on $g \in \mathbb{R}^{2560}$ itself, after \emph{pack centring}. A \emph{pack} is one physical training batch. For the set $S$ of readouts of one view type (such as source or target) in a pack, we subtract the detached mean and normalise:
\begin{equation}
  c(x)_i = \frac{x_i - \sg[\mu_S]}{\norm{x_i - \sg[\mu_S]}}, \qquad \mu_S = \frac{1}{|S|} \sum_{j \in S} x_j . \label{eq:center}
\end{equation}
The shared direction of \cref{tab:geometry} therefore cannot contribute to any loss.

\begin{figure}[t]
  \centering
  \includegraphics[width=\linewidth]{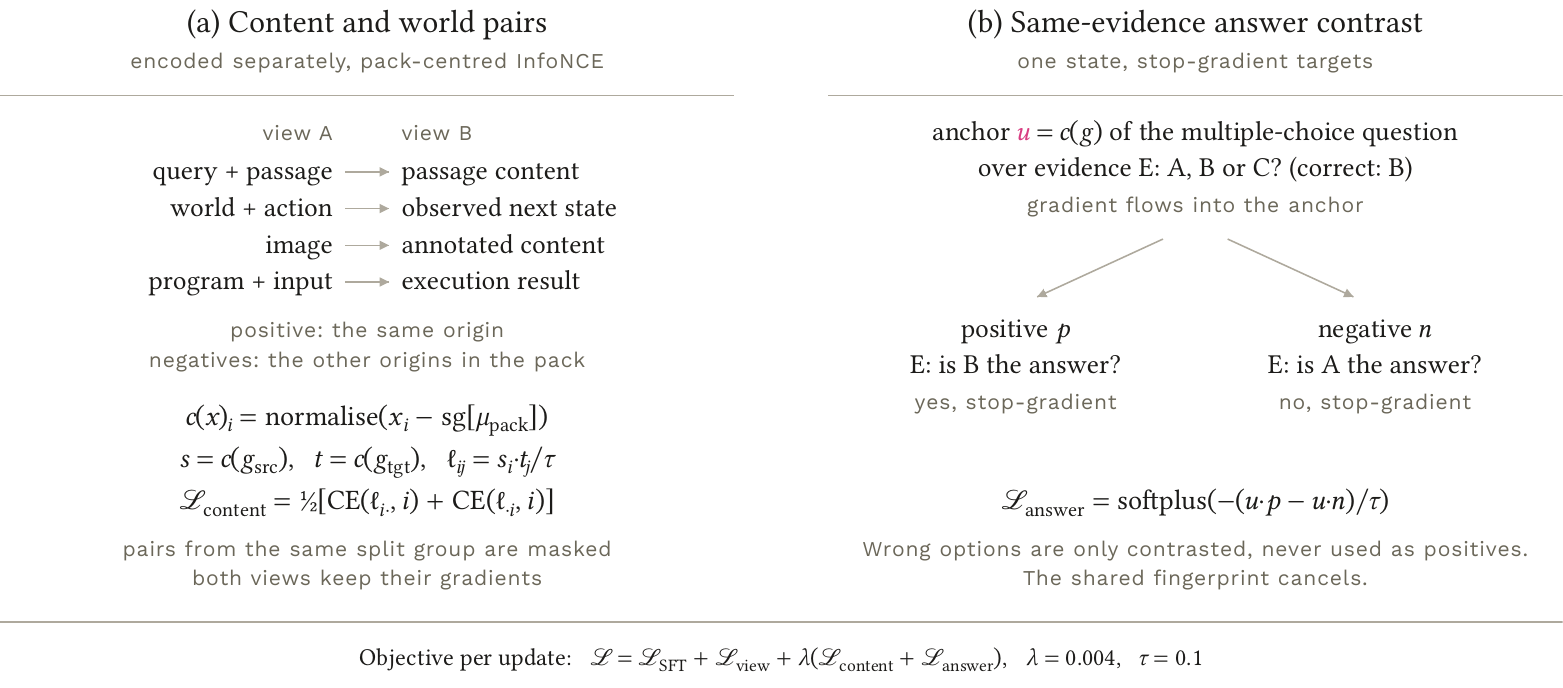}
  \caption{JEPA without a next-token anchor: the stage-2 objective. (a) Separately encoded content and world pairs are centred per pack and trained with a symmetric InfoNCE loss against the other origins in the pack. (b) The anchor is the decision readout of a multiple-choice question. It is contrasted with stop-gradient readouts of a correct ``yes'' event and a wrong ``no'' event under the same evidence, so their shared fingerprint cancels.}
  \label{fig:jepa}
\end{figure}

\textbf{Content pairs} (\cref{fig:jepa}a) use a symmetric InfoNCE loss \citep{oord2018cpc} over the pack, with $s = c(g_{\text{src}})$, $t = c(g_{\text{tgt}})$ and temperature $\tau = 0.1$:
\begin{equation}
  \mathcal{L}_{\text{content}} = \frac{1}{|\mathcal{C}|} \sum_{i \in \mathcal{C}} \tfrac{1}{2} \left[ \CE(\ell_{i\cdot}, i) + \CE(\ell_{\cdot i}, i) \right], \qquad \ell_{ij} = s_i \cdot t_j / \tau, \label{eq:infonce}
\end{equation}
where $\mathcal{C}$ is the set of content pairs in the pack. A split group holds the records derived from one source item, and pairs from the same split group are excluded from each other's negatives. Both views keep their gradients. The other origins in the pack provide the uniformity that pure alignment lacks \citep{wang2020alignment,gao2021simcse}, so the objective rewards correspondence rather than collinearity.

\textbf{Answer states} (\cref{fig:jepa}b) contrast the anchor $u = c(g_{\text{MCQ}})$, the centred decision readout of the multiple-choice question, with two stop-gradient targets under the same evidence. The target $p$ is the readout of the correct option's ``yes'' event, and $n$ is that of a random wrong option's ``no'' event. The loss over the set $\mathcal{A}$ of answer-state origins in the pack is
\begin{equation}
  \mathcal{L}_{\text{answer}} = \frac{1}{|\mathcal{A}|} \sum_{i \in \mathcal{A}} \softplus\!\left( -\frac{u_i \cdot p_i - u_i \cdot n_i}{\tau} \right). \label{eq:answer}
\end{equation}

The event questions carry privileged information, the pointer to the answer, so they serve as targets rather than trainable views. Their readouts are centred separately to remove the global offset between yes and no. Because $p$ and $n$ share evidence and template, their common fingerprint cancels in $u_i \cdot (p_i - n_i)$, and only the direction that separates the correct option from a wrong one remains. Wrong options are only ever contrasted and never used as positives. View supervision gives yes and no events equal weight, so the alternative template carries no label prior.

\textbf{Weighting.} We measured the ratio of gradient norms between the alignment terms and the judgment loss on real packs. At $\lambda = 1$ its median was 128.6, so we set $\lambda = 0.004$, which makes the alignment gradient about half the judgment gradient.

Stage~2 assigns one relation to each origin and replays stage-1 tasks to limit drift. Before the full run, we trained an independent short control, which \cref{sec:jepa-results} reports with the results.

\subsubsection{Results}\label{sec:jepa-results}

We evaluate stage~2 in three ways (\cref{tab:jepa-results}). A representation probe uses held-out origins that never appeared in a stage-2 pair. A frozen panel holds 4,096 origins and 14,606 held-out re-expressions of their questions, such as restatements, logical complements, conditional events, ordinal thresholds and reordered options. Evidence panels test whether judgments change when facts change.

\begin{table}[t]
  \centering
  \caption{Effect of stage~2. The differences combine the alignment terms, the alternative views and continued training. Retrieval is measured within one source, so matching on source fingerprints alone cannot raise the score.}
  \label{tab:jepa-results}
  \small
  \begin{tabular}{@{}lrr@{}}
    \toprule
    Measure & Stage 1 & Stage 2 \\
    \midrule
    \multicolumn{3}{@{}l}{\textit{Within-source retrieval top-1 on never-paired origins} $\uparrow$} \\
    Query $\to$ passage & 15\% & 98\% \\
    Question $\to$ answer content & 4\% & 86\% \\
    Image $\to$ content & 7\% & 67\% \\
    World prediction / causal prediction & 4\% / 1\% & 35\% / 31\% \\
    Answer state: correct event nearer than wrong event & 23\% & 77\% \\
    \midrule
    \multicolumn{3}{@{}l}{\textit{Frozen panel: 14,606 held-out re-expressions of 4,096 questions}} \\
    Accuracy $\uparrow$ & 75.7\% & 85.9\% \\
    NLL $\downarrow$ / Brier $\downarrow$ & 0.613 / 0.342 & 0.437 / 0.217 \\
    Agreement with the original question $\uparrow$ & 78.9\% & 89.0\% \\
    Logical complements answered correctly $\uparrow$ & 44.8\% & 71.1\% \\
    \midrule
    \multicolumn{3}{@{}l}{\textit{Evidence panels: both variants correct after a fact change} $\uparrow$} \\
    Unseen rule worlds / image evidence & 97.7\% / 98.8\% & 98.3\% / 100\% \\
    \bottomrule
  \end{tabular}
\end{table}

\textbf{Representation learning.} For question--answer, query--passage and image content, retrieval of never-paired content rose from single digits or low teens to 67--98\%. The families for world prediction, causal prediction, and assignment and matching also improved. The raw cosine between unrelated answer-state readouts fell from 0.993 to 0.48, so the anisotropy that motivated the objective is largely removed.

\textbf{Judgment consistency.} On held-out re-expressions, accuracy rose by ten points and NLL fell by 29\%. The model improved markedly on logical complements and on both accepting correct options and rejecting wrong ones (\cref{app:jepa}). Evidence sensitivity improved slightly. External calibration, however, moved towards overconfidence: expected calibration error rose from 0.12 to 0.15, and temperature fitting targets this shift (\cref{sec:serving}).

\textbf{Attribution.} Stage~2 combines alternative views, continued training and the alignment terms. An independent control isolates the alignment terms: two 400-update runs started from the stage-1 checkpoint, saw identical batches and differed only in $\lambda$ (0 or 0.004). Alignment raised within-source retrieval of held-out question--answer content from 2.3\% to 56.6\% and of image content from 1.6\% to 39.1\%.

\subsection{Stage 3 (sandbox RL): learning from consequences}\label{sec:rl}

Calibrated decision-making requires feedback grounded in environment transitions rather than static reward signals. Stage~3 therefore trains the model on the outcome distributions of candidate actions, estimated from rollouts or computed by exact oracles. TypeSafe states that RLCD, the training method of Jev, optimises calibrated probabilities \citep{typesafe2026systemone}, and our stage aims at the same effect with its own recipe. In sandbox environments, the model answers outcome questions about candidate actions, takes the action with the highest expected utility and then learns from the observed outcomes.

\textbf{Environments.} The sandbox covers games, grid worlds, verifiable judgments, business simulators, computer use, classification and executable generators (\cref{app:envs}). Classification tasks include structured-record prediction and assistant-intervention decisions.

The model judges \emph{roots}, the sandbox states about which it is questioned. Executable tasks supply exact labels for common deployment judgments. Because fast decision models already execute the steps of browser and mobile agents \citep{jevultrafast2026,zhang2026jevmobile}, computer-use questions ask about task progress, errors, risks and candidate actions over accessibility snapshots \citep{playwright}.

\textbf{Labels and objective.} We fork every legal action from a root and label it with an outcome distribution. The distribution is either a rollout frequency under the continuation policy stated in the question or the exact result of an oracle, such as dynamic programming or an endgame solver. We train on it directly with a proper scoring rule, the cross-entropy against the full outcome distribution. The judgment model is the only learned component, with no value network or reward model.

\textbf{When a policy gradient is calibrated.} RL usually trains with a policy gradient, and the reward decides whether that gradient is calibrated. Consider a question with model distribution $q$ and an outcome distribution $p$ that stays fixed during the update. We sample a predicted outcome $A \sim q$, which the environment never executes, and score it with $C = \mathbf{1}[A = Y]$, where $Y \sim p$. With the reward $R = C - q_A$ held constant, the policy-gradient (score-function) estimator \citep{williams1992reinforce} satisfies
\begin{equation}
  \mathbb{E}\left[ (C - q_A)\, \nabla_\theta \log q_A \right] = \sum_a (p_a - q_a)\, \nabla_\theta q_a = -\tfrac{1}{2}\, \nabla_\theta \norm{q - p}_2^2 . \label{eq:cpg}
\end{equation}
This expectation is half the negative Brier gradient \citep{brier1950verification}, so the fixed point is the true outcome probability rather than certainty. A correctness-only reward, or normalisation by the group standard deviation, instead pushes the top class towards one.

For a fixed observed $Y$, however, the expectation of the estimator over $A$ is exactly the negative gradient of the single-outcome Brier loss $L(q, Y) = \tfrac{1}{2} \sum_k (q_k - \mathbf{1}[Y = k])^2$. The direct loss therefore has the expected gradient of \cref{eq:cpg} without the extra sampling noise. Sampling $A$ helps only when feedback reveals no more than whether a sampled answer was correct. Our environments reveal the outcome itself, so we train with direct proper-scoring losses. \Cref{sec:stage3-eval} reports calibration before and after temperature fitting.

\textbf{Warm start.} Stage~3 opens with supervised updates that introduce question-aware encoding (\cref{sec:parallel}). This warm start combines replayed supervised records, fixed sandbox samples and additional decision questions, including routing, multi-question workflows and larger choice sets.

\textbf{Rounds and curriculum.} Sandbox rounds follow the warm start. Each round plays episodes with the current model, labels legal actions at new roots and measures skill on these fresh roots before any update. Updates combine recent roots with replayed supervised records. Environment sampling follows measured skill: tasks at intermediate skill receive more weight, while every environment retains a share.

\textbf{Selection.} A frozen panel of 16,992 questions at 1,953 roots tracks the stage every two rounds. Panel accuracy rose from 50.6\% to 64.8\%, and the Brier score fell from 0.482 to 0.289. Nearly all of this gain came by round~8. The final model is therefore the uniform average of the five checkpoints from rounds 8 to 15 \citep{izmailov2018swa,wortsman2022soups}. On evaluation sets excluded from stage~3 training, the average matched or beat the individual checkpoints on most measures.

\section{Full-parameter post-training on one GPU}\label{sec:infra}

Every stage trains all 7.09 billion trainable parameters on a single GPU. Stages 1 and 2 ran on an NVIDIA B200 (183~GB) with the stack in \cref{tab:stack} (details in \cref{app:systems}) and took about 51 and 23.4 hours.

Stage~3 ran on a B300 (288~GB). Its warm start used NVFP4 Transformer Engine linear layers for the feed-forward projections, BF16 attention projections and packs of 49,152 tokens, with the rest of the stack unchanged, and took 8.5 hours. The sandbox rounds trained in BF16 and took 2.5 hours in total.

\begin{table}[t]
  \centering
  \caption{Training stack of stages 1 and 2 on a single B200 (PyTorch 2.14, CUDA 13.0).}
  \label{tab:stack}
  \small
  \begin{tabularx}{\linewidth}{@{}P{3cm}L@{}}
    \toprule
    Component & Implementation \\
    \midrule
    Feed-forward blocks & 68 fused Transformer Engine blocks (RMSNorm, gate and up projections, GELU gating, down projection) with NVFP4 matrix multiplications \citep{transformerengine,nvidia2025nvfp4} \\
    Attention projections & 272 projections as rowwise-scaled FP8 matrix multiplications \citep{micikevicius2022fp8,torchao} \\
    Attention & FlashAttention-4 in variable-length mode \citep{zadouri2026fa4,dao2022flashattention}, FlexAttention for sliding-window sequences \citep{dong2024flexattention}, BF16 arithmetic \\
    Storage and optimiser & BF16 parameters and gradients with no FP32 master copy, and 8-bit AdamW \citep{dettmers2022optimizers} with stochastic rounding \citep{gupta2015limited} \\
    Batching and memory & Unpadded packs of 24,576--28,672 tokens (131,072 tokens per update), state keys and values stored once per record, and recomputation \citep{chen2016checkpointing} only for records above the pack limit \\
    Compilation & \texttt{torch.compile} regions \citep{ansel2024pytorch2}, and quantised weights reused until the parameters change \\
    \bottomrule
  \end{tabularx}
\end{table}

Three choices account for most of the efficiency.
\begin{itemize}[beginpenalty=10000]
  \item \textbf{NVFP4 feed-forward blocks.} These blocks hold most of the parameters and most of the matrix work. At 8,192 tokens, their down-projection kernels run 2.1 times faster than in FP8 (0.75 against 1.56~ms). The fused block reproduces the function of the original module.
  \item \textbf{Merged attention in one FlashAttention-4 call.} The keys of each question are laid out as its state followed by its own prefix. The bottom-right-aligned causal mask of the kernel therefore yields the joint softmax of self- and cross-attention. No second pass or log-sum-exp merge is necessary.
  \item \textbf{Unpadded packing.} Projections and feed-forward blocks run over the whole pack as single matrix multiplications, and only attention separates the sequences.
\end{itemize}
A steady-state profile shows that the low-precision matrix multiplications run at about 2.9~PFLOP/s but take only a quarter of GPU kernel time. Fusion and weight reuse target quantisation, transposes, element-wise work and launch overhead.

\section{Serving and calibration}\label{sec:serving}

The command \texttt{bongard serve} implements the System One endpoints of \cref{sec:task}. We fit one temperature per primitive, and one per option count where enough held-out questions exist \citep{guo2017calibration}. The fitting pool resembles deployment: external development sets, exact-probability questions and teacher-authored held-out questions. We bind the temperatures to the checkpoint hash. For the final model, the fitted temperatures lie between 1.2 and 2.15.

Because $h_i$ sees only candidate $i$ and the candidates before it, Choice probabilities are not guaranteed to be invariant to option order. Training reshuffles the options for reviewed templates. An optional serving mode, off by default, also averages each Choice question over cyclic rotations of its options within one request.

The model also runs on a single 24~GB consumer GPU: on an RTX 4090, JevBench's own client measures a median latency of 116~ms per public item.

The same checkpoints run on Apple Silicon, where optional Metal kernels fuse attention as well as query--key normalisation with rotary encoding. Their design draws on MLX \citep{mlx2023} and Metal FlashAttention \citep{turner2024mfa}.

\section{Evaluation}\label{sec:eval}

We evaluate the final model, the uniform average of the stage-3 checkpoints from rounds 8 to 15, with its serving temperatures. Unless noted, timings use one NVIDIA RTX PRO 6000 in BF16. Numbers for other systems come from the public leaderboards, reports and dataset cards named in each table, except our own Jev runs in \cref{sec:same-items}.

The experiments in this section complement the stage-level results of \cref{sec:jepa-results,sec:rl}: they measure the judgment quality of the final model across tasks and the cost of repeated judgments over shared evidence.

\subsection{System One benchmarks}\label{sec:bench}

\begin{table}[tp]
  \centering
  \caption{Public System One benchmarks. DecisionBench, typed-decisions and ImajevBench references come from their leaderboards and dataset cards on 29 September 2026, and JevBench references from its v1.4.2.2 results. On typed-decisions, the gold is a teacher distribution, so the scores measure agreement with that teacher.}
  \label{tab:bench}
  \small
  \begin{tabularx}{\linewidth}{@{}P{2.6cm}P{3.4cm}P{2.4cm}L@{}}
    \toprule
    Benchmark & Measure & Bongard & Reference systems \\
    \midrule
    DecisionBench 1.0 \citep{decisionbench2026} & accuracy, all 23,900 rows & \textbf{78.05\%}\newline (4th of 61) & Imajev-4B 79.7\%, Winnow-12B 76.7\%, Jev 1.13 72.0\%, DeepSeek V4.1 Flash 71.0\%, GPT-5.6 Luna 69.9\% \\
    & ECE and NLL & 0.063 and 0.72 & Imajev-4B 0.069 and 0.69, Jev 1.13 0.128 and 2.43 \\
    JevBench v1.4 \citep{jevbench2026}, public items & accuracy, easy / standard & 100\% / 97.2\% & Imajev-4B 100\% / 99.0\% \\
    typed-decisions \citep{typeddecisions2026}, test & KL from gold, Brier & 0.256, 0.132 & Jev 1.13 1.442, 0.148 \\
    & accuracy & 0.594 & Jev 1.13 0.727, Jeff-Gemma4-E2B 0.561 \\
    behavior-benchmark \citep{behaviorbench2026} & $F_1$ of \emph{present}, core / multilingual & 0.458 / 0.668 & no public results \\
    ImajevBench v2.0-lite \citep{imajevbench2026} & accuracy, dev and calibration splits & 66.9\% (chance 26.5\%) & test split, submitted \\
    \bottomrule
  \end{tabularx}
\end{table}

\Cref{tab:bench} summarises the public benchmarks. With the official DecisionBench runner, Bongard answers 78.05\% of all 23,900 rows correctly and ranks fourth of 61 systems, behind only the benchmark authors' own models and Imajev-4B. It ranks ahead of Jev, Winnow-12B and frontier language models such as GPT-5.6 Luna and DeepSeek V4.1 Flash. Its probabilities are also substantially more reliable than Jev's, with an ECE of 0.063 against 0.128 under the leaderboard's 15 bins and an NLL of 0.72 against 2.43.

On typed-decisions, Bongard's distributions are closer to the teacher's than Jev's are, with a KL of 0.256 against 1.442 and a Brier score of 0.132 against 0.148. Jev agrees with the teacher's top label more often, at 72.7\% against Bongard's 59.4\%. The two kinds of measure capture different properties of a judgment: accuracy concerns only the top-ranked option, whereas KL and Brier score compare the full distribution with the target. ImajevBench asks for decisions from photos and rules, and with the photo withheld, Bongard answers ``unknown'' on 102 of the 103 visual items instead of guessing.

\subsection{Same-item comparison with Jev}\label{sec:same-items}

\begin{figure}[tp]
  \centering
  \includegraphics[width=\linewidth]{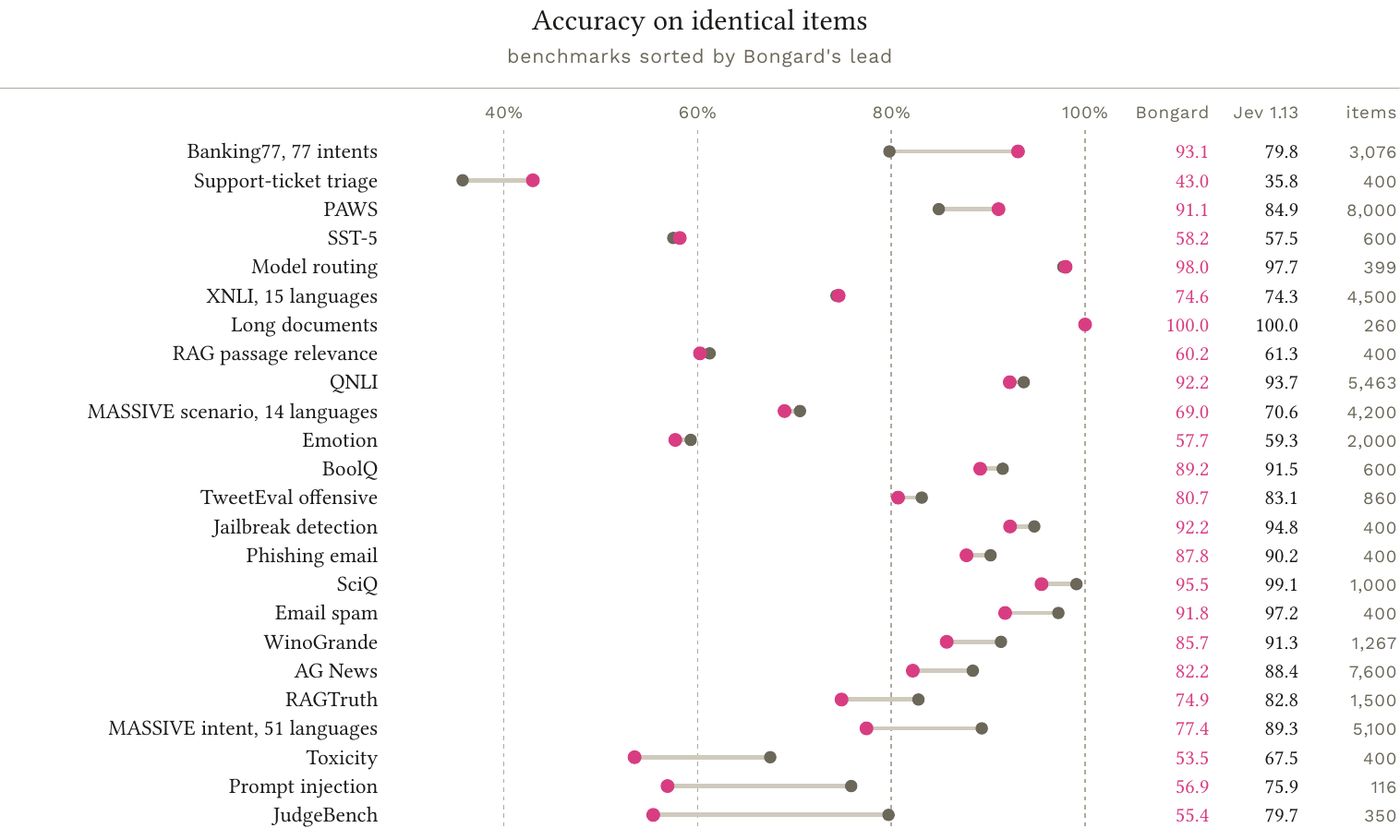}
  \caption{Same-item comparison with Jev~1.13. Both models answered the same 49,291 items from 24 public benchmarks, with identical states, instructions and candidates. Jev answered through TypeSafe's API on 30 September 2026. Dots mark accuracy, and the benchmarks are sorted by Bongard's lead. \Cref{tab:items} lists the item sets.}
  \label{fig:same-items}
\end{figure}

Because leaderboards compare systems on different samples and settings, we sent Bongard's requests unchanged to Jev~1.13 through TypeSafe's API on 30 September 2026 (\cref{fig:same-items}). Most item sets come from the published benchmark code of Laya and Jeff \citep{laya2026code,jeff2026}, and the others are complete public splits (\cref{tab:items}). Both models answered every item. Our API runs reproduce four independent Jev measurements within one point \citep{bakhta2026jevbenchmarks,nibzard2026dmb}.

The open 4B-4B Bongard leads the hosted Jev on six benchmarks and stays within three points on nine more. Its largest lead is on Banking77, where all 77 intents are candidates: 93.1\% against 79.8\%. It also leads on PAWS, with 91.1\% against 84.9\%, and on support-ticket triage.

On support-ticket triage, emotion and SST-5, whose labels are subjective, Jev is overconfident, with an ECE of 0.481, 0.277 and 0.179 against 0.077, 0.032 and 0.051 for Bongard.

\subsection{Languages and long documents}\label{sec:languages}

Laya publishes results across 51 languages and for documents of up to 8,192 tokens, and we rebuilt its items from its code \citep{laya2026,laya2026code}. In its MASSIVE intent sweep, each question offers 20 candidate intents, and 100 questions cover each language \citep{fitzgerald2023massive}. Bongard outperforms both Laya checkpoints in every language. Its mean accuracy is 77.4\%, against 36.6\% for Laya's multilingual checkpoint and 22.7\% for its English one. Its weakest language, Welsh, still reaches 48\%.

Laya's long-document test places a support request after up to 7,000 tokens of meeting notes, and we extend it to 30,000 tokens. The model must assign each request to a department. Bongard assigns all 260 requests correctly, at every length. Laya's multilingual checkpoint reads at most 8,192 tokens, and it assigns 123 of the 160 requests up to 7,000 tokens correctly.

\subsection{Comparison with the pretrained backbone}\label{sec:backbone-eval}

The T5Gemma~2 report scores the pretrained 4B-4B model on three tasks that we also evaluate \citep{zhang2025t5gemma2}, so these tasks show how post-training converts the knowledge of the backbone into direct judgments. Bongard answers each question in one pass of its judgment head, with no examples in the request. It reaches 89.2\% on BoolQ against 79.3\%, and 60.9\% on SocialIQA against 49.9\% \citep{clark2019boolq,sap2019socialiqa}. On WinoGrande, it reaches 85.7\% against 71.6\% for the backbone with five examples \citep{sakaguchi2020winogrande}.

\subsection{Fair random choices}\label{sec:fair}

\begin{figure}[tp]
  \centering
  \includegraphics[width=\linewidth]{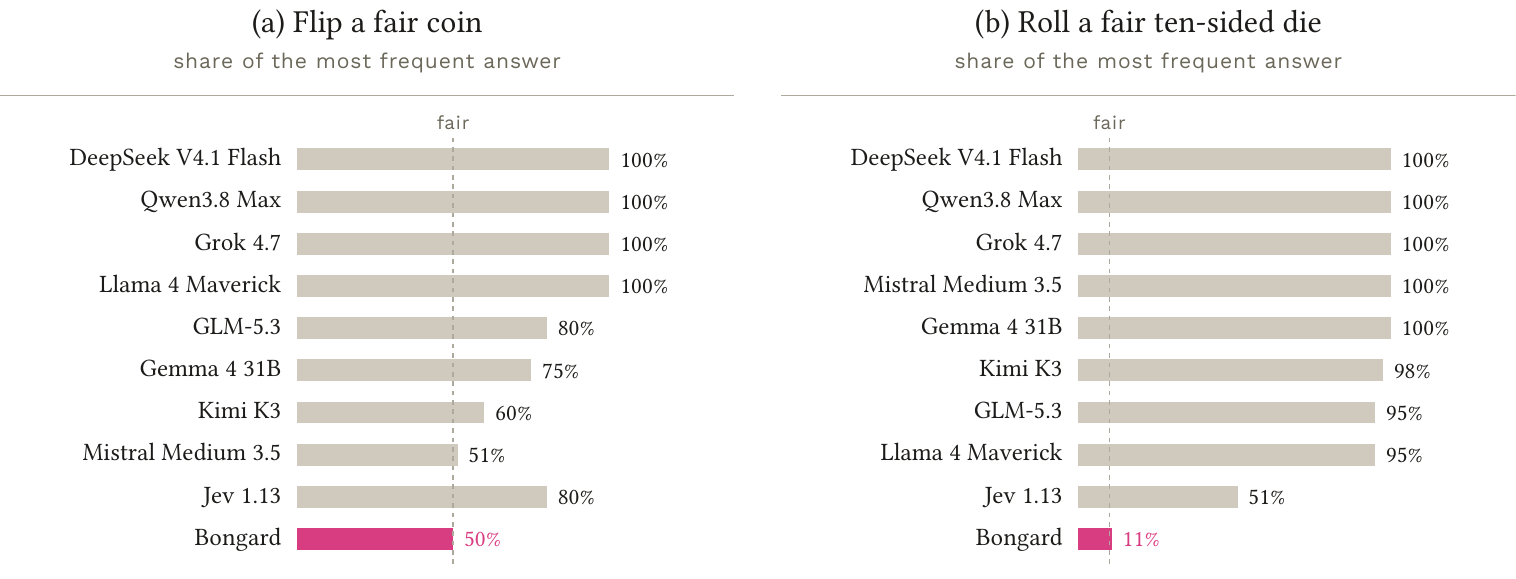}
  \caption{Asked for a fair random choice, chat models repeat one answer, while Bongard returns the uniform distribution. Each chat model answered each question 40 times at its default temperature. For Jev and Bongard, the bar is the largest returned probability. The dashed line marks a fair choice.}
  \label{fig:fair}
\end{figure}

A decision model should also recognise when the evidence favours no option. We asked eight chat models 18 fair-chance questions, such as a coin flip, a die roll, a card suit and a tie-break, 40 times each (\cref{fig:fair}). DeepSeek V4.1 Flash, Qwen3.8 Max, Grok 4.7 and Llama 4 Maverick answered ``heads'' 40 times out of 40. On a ten-sided die, seven of the eight models answered ``7'' in 95--100\% of trials. Jev gives ``heads'' a probability of 0.80. Bongard returns 0.500 for each side, and its mean KL from the uniform distribution over all 18 questions is 0.0009, against 0.769 for Jev.

\subsection{Speed and cost}\label{sec:speed}

\begin{figure}[tp]
  \centering
  \includegraphics[width=\linewidth]{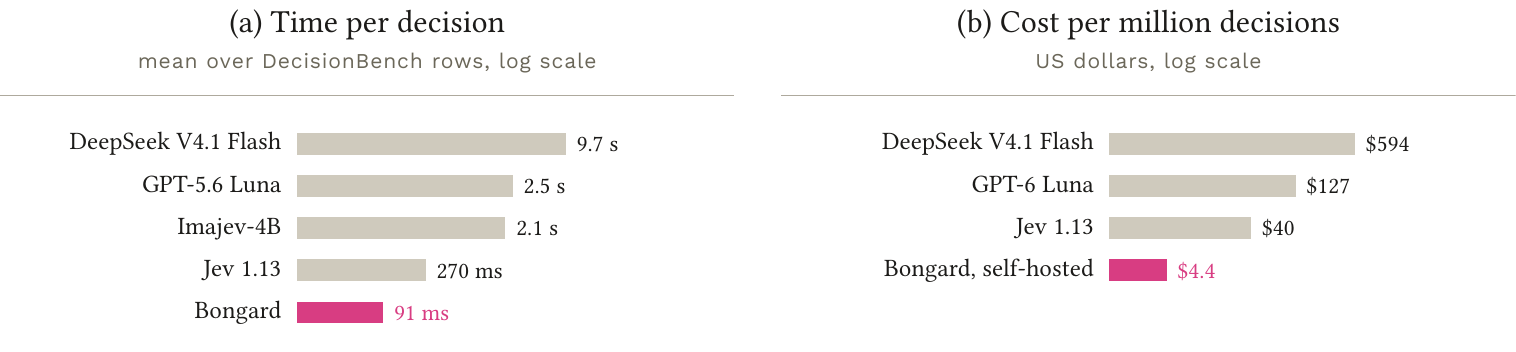}
  \caption{Time and cost per decision. Bongard runs on one RTX PRO 6000 in BF16, and its cost assumes \$1.79 per GPU hour. Latencies of the other systems come from the DecisionBench leaderboard and include network time. Costs come from the JevBench results, as dollars per 1,000 decisions times 1,000.}
  \label{fig:speed}
\end{figure}

\begin{figure}[tp]
  \centering
  \includegraphics[width=\linewidth]{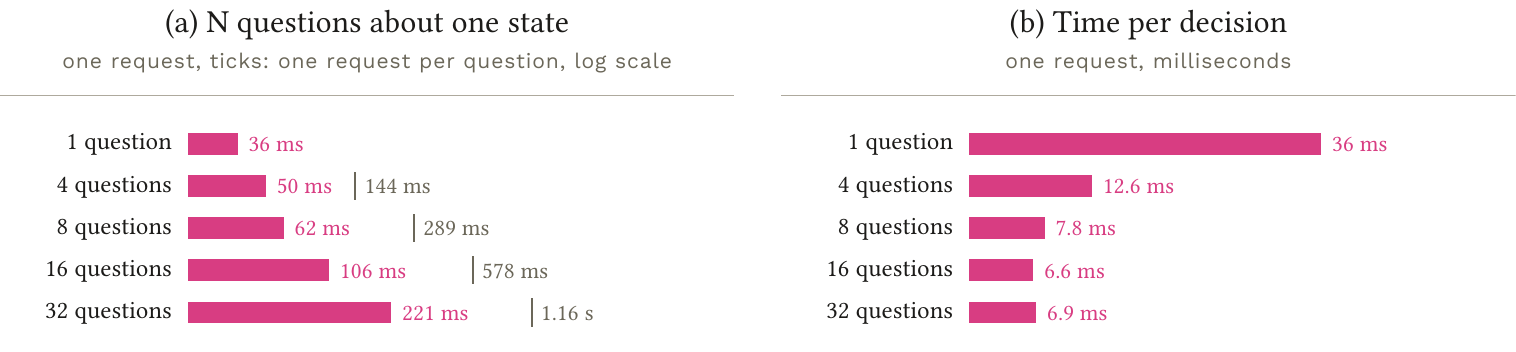}
  \caption{Latency of shared-state requests. One request asks $N$ questions about the same JevBench state, and the encoder reads the state once. Ticks mark $N$ separate single-question requests, at $N$ times the single-question latency. Thirty-two questions take 221~ms in one request instead of 1.16~s, so the time per decision falls from 36 to 6.9~ms. One RTX PRO 6000, BF16.}
  \label{fig:shared}
\end{figure}

One request at a time, the median latency is 36~ms on short JevBench items, and the mean over DecisionBench rows is 91~ms (\cref{fig:speed}). Questions about the same state share one encoder pass (\cref{fig:shared}). Thirty-two such questions take 221~ms in one request, 5.2 times less than as separate requests, or 6.9~ms per decision. Batched requests reach about 400,000 decisions per hour on one GPU. At \$1.79 per GPU hour, a million decisions cost about \$4.4. Jev costs \$40 and GPT-6 Luna \$127 for the same number.

\subsection{Decisions in games}\label{sec:games}

\begin{table}[tp]
  \centering
  \caption{Games played by Bongard. The game engine lists the legal moves and states the facts of each move in the state. The question gives a one-line strategy in plain English, and the model chooses every move. For 2048 and Tetris, the facts include the engine's rating of each move. Doom is ViZDoom Defend the Center \citep{kempka2016vizdoom}, and the model reads a one-line description of the scene every 0.2~s.}
  \label{tab:games}
  \small
  \begin{tabularx}{\linewidth}{@{}P{2.4cm}P{2.2cm}L@{}}
    \toprule
    Game & Games & Result \\
    \midrule
    2048 \citep{cirulli2014twenty48} & 12 & reaches the 2048 tile in 9 games and 4096 in 2, best score 55,140 \\
    Tetris & 5 & never tops out within the limit of 1,500 pieces, 596--599 rows cleared per game \\
    Othello & 32 + 32 & wins 24 of 32 against a greedy engine and 4 of 32 against a two-ply positional engine, one of them 13--0 in nine moves \\
    Minesweeper, 8$\times$8 & 64 & clears the board in 42 games \\
    Snake, 8$\times$8 & 32 & grows to 32 cells, half of the board \\
    Doom & 32 & 10.3 kills per 30-second episode on average, 20 at best \\
    \bottomrule
  \end{tabularx}
\end{table}

Games test decisions that must follow each other quickly (\cref{tab:games}). In each board game, the engine computes objective facts about each legal move, such as the score, the free cells and the replies it allows. It writes them into the state, and a single sentence states the strategy. Bongard then chooses the move, so the same model can play a new game from a new sentence. In 2048 it reached the 2048 tile in 9 of 12 games. In Tetris it never topped out in 1,500 pieces.

\subsection{What stage 3 adds}\label{sec:stage3-eval}

Stage~3 raises accuracy on every held-out set in \cref{tab:stage3}. The largest gains are on hard-style decisions, from 34.9\% to 50.5\%, and on frontier-style decisions, from 45.2\% to 73.9\%. Fair-chance questions become almost exactly uniform. Serving temperatures improve calibration further. On the external development sets, NLL falls from 0.824 to 0.691 and ECE from 0.141 to 0.030.

\begin{table}[htbp]
  \centering
  \caption{Stage 2 (JEPA) and the final model on evaluation sets excluded from stage~3 training. The style-based sets are internal panels, and the public-benchmark samples cover multiple tasks.}
  \label{tab:stage3}
  \small
  \begin{tabularx}{\linewidth}{@{}Lrr@{}}
    \toprule
    Evaluation set & Stage 2 & Final \\
    \midrule
    Hard-style decisions (109) & 34.9\% & 50.5\% \\
    Sealed-style decisions (200) & 35.0\% & 50.5\% \\
    Frontier-style decisions (199) & 45.2\% & 73.9\% \\
    DecisionBench, balanced sample (2,580) & 64.3\% & 68.7\% \\
    RouterBench routing between cheap and strong models (600) & 56.2\% & 62.5\% \\
    JevBench public standard items (72) & 93.1\% & 97.2\% \\
    Fair-chance KL from uniform (18 questions) & 0.173 & 0.0009 \\
    \bottomrule
  \end{tabularx}
\end{table}

\FloatBarrier
\section{Related work}\label{sec:related}

\textbf{System One models.} TypeSafe introduced System One models, Jev and its typed API \citep{typesafe2026systemone,typesafe2026docs}. From the behaviour of that API, \citet{hume2026jev} infers a decoder-only backbone with prefix caching. \citet{ling2026jevwild} analyse 2,170 public projects that use Jev as a reusable decision component. \citet{zhang2026jevmobile} pairs a planning vision-language model with Jev as a fast executor of mobile GUI actions. This design cuts execution time and cost at a small loss in success rate. Further open systems behind the System One API include the Open-Jev project of \citet{cai2026openjev}, Cygnet \citep{cygnet2026} and decider \citep{decider2026}, and JevBench compares many of them \citep{jevbench2026}.

\textbf{Calibration and single-pass reasoning.} Decision calibration asks that probabilities be reliable for the decisions that they drive \citep{zhao2021decisioncal}. Proper-scoring rewards train the verbalised confidence of generative models \citep{banihirouni2025doubt,damani2025rlcr}, and community RLCD-style projects use them too \citep{everlcd2026,nanojev2026}. Beyond chess and Othello (\cref{sec:intro}), single-pass transformers can also internalise step-by-step reasoning \citep{deng2024implicitcot}.

\section{Limitations}\label{sec:limits}

The alignment terms are isolated only by a short independent control, which shows a reshaped representation but not yet a decision gain. The head's subspace also captures less of $g$ than before, so a full-length control with $\lambda = 0$ and a readout that uses the reshaped representation come next. Stage~3 tracks its progress on a panel from its own environments. Evaluation sets excluded from stage~3 training improve as well (\cref{tab:stage3}), but judgment of consequences in unseen environments is not yet tested directly. Verifying worked answers is less reliable than other judgments, because the model tends to accept a wrong answer as correct.

Jev still leads on several kinds of judgment. On typed-decisions, it agrees with the teacher's top label more often. On identical items, it leads by more than three points on 9 of 24 public benchmarks, most on JudgeBench, prompt injection and toxicity.

The stage-1 and stage-2 checkpoints encode the state alone, so their encoder cannot focus on what a question concerns. Question-aware encoding removes this limit at a cost. An answer can change when other instructions join the request, and the encoder cache serves only requests with the same instructions.

\section{Conclusion}\label{sec:conclusion}

We present Bongard, an open System One model that formulates machine intuition as direct, callable probabilistic judgment. It combines an encoder--decoder backbone, which separates the reading of a state from the judgments made about it, with joint-embedding representation learning and sandbox outcome feedback. Each of its three training stages updates all trainable parameters on a single GPU.

Joint-embedding training reshapes the decision representation, and sandbox training improves outcome judgments and held-out task scores. The final model reaches 78.05\% accuracy on DecisionBench and answers 32 questions about one state in 221~ms. The released weights and runtime, together with the training method described here, provide a concrete route to machine intuition. Further work will examine how to use the learned representation more fully and how to allocate capacity between the encoder and the decoder.

\bibliographystyle{plainnat}
\bibliography{refs}

@book{kahneman2011thinking,
  author    = {Kahneman, Daniel},
  title     = {{Thinking, Fast and Slow}},
  publisher = {Farrar, Straus and Giroux},
  address   = {New York},
  year      = {2011}
}

@article{kahneman2009conditions,
  author  = {Kahneman, Daniel and Klein, Gary},
  title   = {{Conditions for Intuitive Expertise: A Failure to Disagree}},
  journal = {American Psychologist},
  volume  = {64},
  number  = {6},
  pages   = {515--526},
  year    = {2009},
  doi     = {10.1037/a0016755}
}

@article{wilson1991thinking,
  author  = {Wilson, Timothy D. and Schooler, Jonathan W.},
  title   = {{Thinking Too Much: Introspection Can Reduce the Quality of Preferences and Decisions}},
  journal = {Journal of Personality and Social Psychology},
  volume  = {60},
  number  = {2},
  pages   = {181--192},
  year    = {1991},
  doi     = {10.1037/0022-3514.60.2.181}
}

@misc{palmer2026kev,
  author       = {Palmer, Jared and {Kev contributors}},
  title        = {{Kev: Jev-like Decision Models Built on Qwen}},
  year         = {2026},
  howpublished = {\url{https://github.com/jaredpalmer/kev}},
  note         = {Implementation and model cards, accessed 30 September 2026}
}

@misc{typesafe2026systemone,
  author       = {{TypeSafe AI}},
  title        = {{Introducing System One Models and Jev}},
  year         = {2026},
  howpublished = {\url{https://typesafe.ai/blog/introducing-system-one-models-and-jev}},
  note         = {Blog post}
}

@misc{typesafe2026docs,
  author       = {{TypeSafe AI}},
  title        = {{System One API Documentation: Primitives and Parallel Questions}},
  year         = {2026},
  howpublished = {\url{https://docs.typesafe.ai}},
  note         = {Accessed September 2026}
}

@misc{hume2026jev,
  author       = {Hume, Archer},
  title        = {{Jev's Architecture Unmasked}},
  year         = {2026},
  month        = sep,
  howpublished = {\url{https://archerhume.com/posts/jevs-architecture-unmasked/}},
  note         = {Blog post}
}

@misc{openjev2026,
  author       = {{OpenJev contributors}},
  title        = {{OpenJev: An Open-Source System One Decision Server}},
  year         = {2026},
  howpublished = {\url{https://github.com/razorback16/openjev}}
}

@misc{autojev2026,
  author       = {{denis-pplx}},
  title        = {{AutoJev-27B}},
  year         = {2026},
  howpublished = {\url{https://github.com/denis-pplx/autojev}}
}

@misc{mojev2026,
  author       = {{MoLeMo Lab}},
  title        = {{MoJev}},
  year         = {2026},
  howpublished = {\url{https://huggingface.co/MoLeMo-Lab/mojev}}
}

@misc{clm2026,
  author       = {{Contrastive-LM}},
  title        = {{CLM-v0.1-8B}},
  year         = {2026},
  howpublished = {\url{https://github.com/Contrastive-LM/CLM}}
}

@misc{laya2026,
  author       = {{Convai Innovations}},
  title        = {{Laya: A Typed-Decisions Model}},
  year         = {2026},
  howpublished = {\url{https://huggingface.co/convaiinnovations/laya-typed-decisions}}
}

@misc{jevk52026,
  author       = {{allebee}},
  title        = {{JevK5}},
  year         = {2026},
  howpublished = {\url{https://github.com/allebee/jevk5}}
}

@misc{plumb2026,
  author       = {{crh225}},
  title        = {{Plumb-4B}},
  year         = {2026},
  howpublished = {\url{https://github.com/crh225/plumb}}
}

@misc{decider2026,
  author       = {{Mapika}},
  title        = {{decider}},
  year         = {2026},
  howpublished = {\url{https://github.com/Mapika/decider}}
}

@misc{cygnet2026,
  author       = {{blockbrain}},
  title        = {{Cygnet Recipe}},
  year         = {2026},
  howpublished = {\url{https://github.com/blockbrain-ai/cygnet-recipe}}
}

@misc{cai2026openjev,
  author       = {Cai, Zefan},
  title        = {{Open-Jev}},
  year         = {2026},
  howpublished = {\url{https://github.com/Zefan-Cai/Open-Jev}}
}

@misc{semif2026,
  author       = {{TheoLeeCJ}},
  title        = {{SemIf}},
  year         = {2026},
  howpublished = {\url{https://github.com/TheoLeeCJ/SemIf}}
}

@misc{jevfire2026,
  author       = {{kikoncuo}},
  title        = {{Jevfire}},
  year         = {2026},
  howpublished = {\url{https://github.com/kikoncuo/jevfire}}
}

@misc{openjevsglang2026,
  author       = {{ekzhang}},
  title        = {{openjev-sglang}},
  year         = {2026},
  howpublished = {\url{https://github.com/ekzhang/openjev-sglang}}
}

@misc{everlcd2026,
  author       = {{anthony-maio}},
  title        = {{eve-rlcd}},
  year         = {2026},
  howpublished = {\url{https://github.com/anthony-maio/eve-rlcd}}
}

@misc{nanojev2026,
  author       = {{TianyuCodings}},
  title        = {{NanoJev}},
  year         = {2026},
  howpublished = {\url{https://github.com/TianyuCodings/NanoJev}}
}

@misc{jevultrafast2026,
  author       = {{Browser Use}},
  title        = {{Jev Ultrafast}},
  year         = {2026},
  howpublished = {\url{https://github.com/browser-use/jev-ultrafast}}
}

@article{ling2026jevwild,
  author  = {Ling, Guoming and Xue, Muen and Ye, Zijian},
  title   = {{Jev in the Wild: A Data-Driven Analysis of the Jev Model's Functionality, Applications and Ecosystem}},
  journal = {arXiv preprint arXiv:2609.30216},
  year    = {2026}
}

@article{zhang2026jevmobile,
  author  = {Zhang, Linghua},
  title   = {{Jev-Mobile: Jev as an Executor for Mobile GUI Agents}},
  journal = {arXiv preprint arXiv:2609.30186},
  year    = {2026}
}

@misc{motherduck2026,
  author       = {{MotherDuck}},
  title        = {{Introducing prompt\_jev(): Bringing Jev to MotherDuck SQL}},
  year         = {2026},
  howpublished = {\url{https://motherduck.com/blog/motherduck-supports-jev/}},
  note         = {Blog post}
}

@misc{jevbench2026,
  author       = {Standhartinger, Florian and contributors},
  title        = {{JevBench: A Benchmark for System One Decision Models}},
  year         = {2026},
  howpublished = {\url{https://github.com/fstandhartinger/jevbench}}
}

@misc{typeddecisions2026,
  author       = {{LocalLLaMA}},
  title        = {{typed-decisions}},
  year         = {2026},
  howpublished = {\url{https://huggingface.co/datasets/LocalLLaMA/typed-decisions}}
}

@misc{behaviorbench2026,
  author       = {{Respan AI}},
  title        = {{behavior-benchmark}},
  year         = {2026},
  howpublished = {\url{https://huggingface.co/datasets/respanai/behavior-benchmark}}
}

@book{bongard1970pattern,
  author    = {Bongard, Mikhail M.},
  title     = {{Pattern Recognition}},
  publisher = {Spartan Books},
  address   = {New York},
  year      = {1970}
}

@article{raffel2020t5,
  author  = {Raffel, Colin and Shazeer, Noam and Roberts, Adam and Lee, Katherine and Narang, Sharan and Matena, Michael and Zhou, Yanqi and Li, Wei and Liu, Peter J.},
  title   = {{Exploring the Limits of Transfer Learning with a Unified Text-to-Text Transformer}},
  journal = {Journal of Machine Learning Research},
  volume  = {21},
  number  = {140},
  pages   = {1--67},
  year    = {2020}
}

@inproceedings{tay2023ul2,
  author    = {Tay, Yi and Dehghani, Mostafa and Tran, Vinh Q. and Garcia, Xavier and Wei, Jason and Wang, Xuezhi and Chung, Hyung Won and Bahri, Dara and Schuster, Tal and Zheng, Huaixiu Steven and Zhou, Denny and Houlsby, Neil and Metzler, Donald},
  title     = {{UL2: Unifying Language Learning Paradigms}},
  booktitle = {International Conference on Learning Representations},
  year      = {2023}
}

@article{gemma3,
  author  = {{Gemma Team}},
  title   = {{Gemma 3 Technical Report}},
  journal = {arXiv preprint arXiv:2503.19786},
  year    = {2025}
}

@article{zhang2025encdecgemma,
  author  = {Zhang, Biao and Moiseev, Fedor and Ainslie, Joshua and Suganthan, Paul and Ma, Min and Bhupatiraju, Surya and Lebron, Fede and Firat, Orhan and Joulin, Armand and Dong, Zhe},
  title   = {{Encoder-Decoder Gemma: Improving the Quality-Efficiency Trade-Off via Adaptation}},
  journal = {arXiv preprint arXiv:2504.06225},
  year    = {2025}
}

@article{zhang2025t5gemma2,
  author  = {Zhang, Biao and Suganthan, Paul and Liu, Ga{\"e}l and Philippov, Ilya and Dua, Sahil and Hora, Ben and Black, Kat and Martins, Gus and Sanseviero, Omar and Pathak, Shreya and Hardin, Cassidy and Visin, Francesco and Zhang, Jiageng and Kenealy, Kathleen and Yin, Qin and Lacombe, Olivier and Joulin, Armand and Warkentin, Tris and Roberts, Adam},
  title   = {{T5Gemma 2: Seeing, Reading, and Understanding Longer}},
  journal = {arXiv preprint arXiv:2512.14856},
  year    = {2025}
}

@article{zhang2025redllm,
  author  = {Zhang, Biao and Cheng, Yong and Shakeri, Siamak and Wang, Xinyi and Ma, Min and Firat, Orhan},
  title   = {{Encoder-Decoder or Decoder-Only? Revisiting Encoder-Decoder Large Language Model}},
  journal = {arXiv preprint arXiv:2510.26622},
  year    = {2025}
}

@inproceedings{zhai2023siglip,
  author    = {Zhai, Xiaohua and Mustafa, Basil and Kolesnikov, Alexander and Beyer, Lucas},
  title     = {{Sigmoid Loss for Language Image Pre-Training}},
  booktitle = {Proceedings of the IEEE/CVF International Conference on Computer Vision},
  year      = {2023}
}

@inproceedings{izacard2021fid,
  author    = {Izacard, Gautier and Grave, Edouard},
  title     = {{Leveraging Passage Retrieval with Generative Models for Open Domain Question Answering}},
  booktitle = {Proceedings of the 16th Conference of the European Chapter of the Association for Computational Linguistics},
  year      = {2021}
}

@inproceedings{nogueira2020monot5,
  author    = {Nogueira, Rodrigo and Jiang, Zhiying and Pradeep, Ronak and Lin, Jimmy},
  title     = {{Document Ranking with a Pretrained Sequence-to-Sequence Model}},
  booktitle = {Findings of the Association for Computational Linguistics: EMNLP 2020},
  year      = {2020}
}

@inproceedings{zhuang2023rankt5,
  author    = {Zhuang, Honglei and Qin, Zhen and Jagerman, Rolf and Hui, Kai and Ma, Ji and Lu, Jing and Ni, Jianmo and Wang, Xuanhui and Bendersky, Michael},
  title     = {{RankT5: Fine-Tuning T5 for Text Ranking with Ranking Losses}},
  booktitle = {Proceedings of the 46th International ACM SIGIR Conference on Research and Development in Information Retrieval},
  year      = {2023}
}

@inproceedings{zhao2021calibrate,
  author    = {Zhao, Tony Z. and Wallace, Eric and Feng, Shi and Klein, Dan and Singh, Sameer},
  title     = {{Calibrate Before Use: Improving Few-Shot Performance of Language Models}},
  booktitle = {Proceedings of the 38th International Conference on Machine Learning},
  year      = {2021}
}

@inproceedings{zheng2024mcq,
  author    = {Zheng, Chujie and Zhou, Hao and Meng, Fandong and Zhou, Jie and Huang, Minlie},
  title     = {{Large Language Models Are Not Robust Multiple Choice Selectors}},
  booktitle = {International Conference on Learning Representations},
  year      = {2024}
}

@misc{su2023decoderonly,
  author       = {Su, Jianlin},
  title        = {{Why Are Current LLMs All Decoder-Only Architectures?}},
  year         = {2023},
  howpublished = {\url{https://spaces.ac.cn/archives/9529}},
  note         = {Blog post, Scientific Spaces (in Chinese)}
}

@inproceedings{dong2021rank,
  author    = {Dong, Yihe and Cordonnier, Jean-Baptiste and Loukas, Andreas},
  title     = {{Attention Is Not All You Need: Pure Attention Loses Rank Doubly Exponentially with Depth}},
  booktitle = {Proceedings of the 38th International Conference on Machine Learning},
  year      = {2021}
}

@article{peng2025rwkv7,
  author  = {Peng, Bo and Zhang, Ruichong and Goldstein, Daniel and Alcaide, Eric and Du, Xingjian and Hou, Haowen and Lin, Jiaju and Liu, Jiaxing and Lu, Janna and Merrill, William and Song, Guangyu and Tan, Kaifeng and Utpala, Saiteja and Wilce, Nathan and Wind, Johan S. and Wu, Tianyi and Wuttke, Daniel and Zhou-Zheng, Christian},
  title   = {{RWKV-7 ``Goose'' with Expressive Dynamic State Evolution}},
  journal = {arXiv preprint arXiv:2503.14456},
  year    = {2025}
}

@article{zhou2024valueresidual,
  author  = {Zhou, Zhanchao and Wu, Tianyi and Jiang, Zhiyun and Obeid, Fares and Lan, Zhenzhong},
  title   = {{Value Residual Learning}},
  journal = {arXiv preprint arXiv:2410.17897},
  year    = {2024}
}

@article{juravsky2024hydragen,
  author  = {Juravsky, Jordan and Brown, Bradley and Ehrlich, Ryan and Fu, Daniel Y. and R{\'e}, Christopher and Mirhoseini, Azalia},
  title   = {{Hydragen: High-Throughput LLM Inference with Shared Prefixes}},
  journal = {arXiv preprint arXiv:2402.05099},
  year    = {2024}
}

@misc{lecun2022path,
  author       = {LeCun, Yann},
  title        = {{A Path Towards Autonomous Machine Intelligence}},
  year         = {2022},
  howpublished = {OpenReview preprint, \url{https://openreview.net/forum?id=BZ5a1r-kVsf}}
}

@article{dawid2023lvebm,
  author  = {Dawid, Anna and LeCun, Yann},
  title   = {{Introduction to Latent Variable Energy-Based Models: A Path Towards Autonomous Machine Intelligence}},
  journal = {arXiv preprint arXiv:2306.02572},
  year    = {2023}
}

@inproceedings{assran2023ijepa,
  author    = {Assran, Mahmoud and Duval, Quentin and Misra, Ishan and Bojanowski, Piotr and Vincent, Pascal and Rabbat, Michael and LeCun, Yann and Ballas, Nicolas},
  title     = {{Self-Supervised Learning from Images with a Joint-Embedding Predictive Architecture}},
  booktitle = {Proceedings of the IEEE/CVF Conference on Computer Vision and Pattern Recognition},
  year      = {2023}
}

@article{bardes2024vjepa,
  author  = {Bardes, Adrien and Garrido, Quentin and Ponce, Jean and Chen, Xinlei and Rabbat, Michael and LeCun, Yann and Assran, Mahmoud and Ballas, Nicolas},
  title   = {{Revisiting Feature Prediction for Learning Visual Representations from Video}},
  journal = {arXiv preprint arXiv:2404.08471},
  year    = {2024}
}

@article{assran2025vjepa2,
  author  = {Assran, Mido and others},
  title   = {{V-JEPA 2: Self-Supervised Video Models Enable Understanding, Prediction and Planning}},
  journal = {arXiv preprint arXiv:2506.09985},
  year    = {2025}
}

@article{huang2025llmjepa,
  author  = {Huang, Hai and LeCun, Yann and Balestriero, Randall},
  title   = {{LLM-JEPA: Large Language Models Meet Joint Embedding Predictive Architectures}},
  journal = {arXiv preprint arXiv:2509.14252},
  year    = {2025}
}

@article{gillin2026bertjepa,
  author  = {Gillin, Taj and Lalani, Adam and Zhang, Kenneth and Mateos Salles, Marcel},
  title   = {{BERT-JEPA: Reorganizing CLS Embeddings for Language-Invariant Semantics}},
  journal = {arXiv preprint arXiv:2601.00366},
  year    = {2026}
}

@article{oord2018cpc,
  author  = {van den Oord, A{\"a}ron and Li, Yazhe and Vinyals, Oriol},
  title   = {{Representation Learning with Contrastive Predictive Coding}},
  journal = {arXiv preprint arXiv:1807.03748},
  year    = {2018}
}

@inproceedings{ethayarajh2019contextual,
  author    = {Ethayarajh, Kawin},
  title     = {{How Contextual are Contextualized Word Representations? Comparing the Geometry of BERT, ELMo, and GPT-2 Embeddings}},
  booktitle = {Proceedings of the 2019 Conference on Empirical Methods in Natural Language Processing},
  year      = {2019}
}

@inproceedings{wang2020alignment,
  author    = {Wang, Tongzhou and Isola, Phillip},
  title     = {{Understanding Contrastive Representation Learning through Alignment and Uniformity on the Hypersphere}},
  booktitle = {Proceedings of the 37th International Conference on Machine Learning},
  year      = {2020}
}

@inproceedings{gao2021simcse,
  author    = {Gao, Tianyu and Yao, Xingcheng and Chen, Danqi},
  title     = {{SimCSE: Simple Contrastive Learning of Sentence Embeddings}},
  booktitle = {Proceedings of the 2021 Conference on Empirical Methods in Natural Language Processing},
  year      = {2021}
}

@article{brier1950verification,
  author  = {Brier, Glenn W.},
  title   = {{Verification of Forecasts Expressed in Terms of Probability}},
  journal = {Monthly Weather Review},
  volume  = {78},
  number  = {1},
  pages   = {1--3},
  year    = {1950}
}

@article{gneiting2007strictly,
  author  = {Gneiting, Tilmann and Raftery, Adrian E.},
  title   = {{Strictly Proper Scoring Rules, Prediction, and Estimation}},
  journal = {Journal of the American Statistical Association},
  volume  = {102},
  number  = {477},
  pages   = {359--378},
  year    = {2007}
}

@inproceedings{guo2017calibration,
  author    = {Guo, Chuan and Pleiss, Geoff and Sun, Yu and Weinberger, Kilian Q.},
  title     = {{On Calibration of Modern Neural Networks}},
  booktitle = {Proceedings of the 34th International Conference on Machine Learning},
  year      = {2017}
}

@inproceedings{zhao2021decisioncal,
  author    = {Zhao, Shengjia and Kim, Michael P. and Sahoo, Roshni and Ma, Tengyu and Ermon, Stefano},
  title     = {{Calibrating Predictions to Decisions: A Novel Approach to Multi-Class Calibration}},
  booktitle = {Advances in Neural Information Processing Systems},
  volume    = {34},
  year      = {2021}
}

@article{williams1992reinforce,
  author  = {Williams, Ronald J.},
  title   = {{Simple Statistical Gradient-Following Algorithms for Connectionist Reinforcement Learning}},
  journal = {Machine Learning},
  volume  = {8},
  pages   = {229--256},
  year    = {1992}
}

@article{damani2025rlcr,
  author  = {Damani, Mehul and Puri, Isha and Slocum, Stewart and Shenfeld, Idan and Choshen, Leshem and Kim, Yoon and Andreas, Jacob},
  title   = {{Beyond Binary Rewards: Training LMs to Reason About Their Uncertainty}},
  journal = {arXiv preprint arXiv:2507.16806},
  year    = {2025}
}

@article{banihirouni2025doubt,
  author  = {Bani-Harouni, David and Pellegrini, Chantal and Stangel, Paul and {\"O}zsoy, Ege and Zaripova, Kamilia and Navab, Nassir and Keicher, Matthias},
  title   = {{Rewarding Doubt: A Reinforcement Learning Approach to Calibrated Confidence Expression of Large Language Models}},
  journal = {arXiv preprint arXiv:2503.02623},
  year    = {2025}
}

@article{ruoss2024chess,
  author  = {Ruoss, Anian and Del{\'e}tang, Gr{\'e}goire and Medapati, Sourabh and Grau-Moya, Jordi and Wenliang, Li Kevin and Catt, Elliot and Reid, John and Lewis, Cannada A. and Veness, Joel and Genewein, Tim},
  title   = {{Amortized Planning with Large-Scale Transformers: A Case Study on Chess}},
  journal = {arXiv preprint arXiv:2402.04494},
  year    = {2024}
}

@inproceedings{li2023othello,
  author    = {Li, Kenneth and Hopkins, Aspen K. and Bau, David and Vi{\'e}gas, Fernanda and Pfister, Hanspeter and Wattenberg, Martin},
  title     = {{Emergent World Representations: Exploring a Sequence Model Trained on a Synthetic Task}},
  booktitle = {International Conference on Learning Representations},
  year      = {2023}
}

@article{deng2024implicitcot,
  author  = {Deng, Yuntian and Choi, Yejin and Shieber, Stuart},
  title   = {{From Explicit CoT to Implicit CoT: Learning to Internalize CoT Step by Step}},
  journal = {arXiv preprint arXiv:2405.14838},
  year    = {2024}
}

@misc{playwright,
  author       = {{Microsoft}},
  title        = {{Playwright}},
  year         = {2026},
  howpublished = {\url{https://playwright.dev}}
}

@inproceedings{izmailov2018swa,
  author    = {Izmailov, Pavel and Podoprikhin, Dmitrii and Garipov, Timur and Vetrov, Dmitry and Wilson, Andrew Gordon},
  title     = {{Averaging Weights Leads to Wider Optima and Better Generalization}},
  booktitle = {Proceedings of the 34th Conference on Uncertainty in Artificial Intelligence},
  year      = {2018}
}

@inproceedings{wortsman2022soups,
  author    = {Wortsman, Mitchell and Ilharco, Gabriel and Gadre, Samir Yitzhak and Roelofs, Rebecca and Gontijo-Lopes, Raphael and Morcos, Ari S. and Namkoong, Hongseok and Farhadi, Ali and Carmon, Yair and Kornblith, Simon and Schmidt, Ludwig},
  title     = {{Model Soups: Averaging Weights of Multiple Fine-Tuned Models Improves Accuracy without Increasing Inference Time}},
  booktitle = {Proceedings of the 39th International Conference on Machine Learning},
  year      = {2022}
}

@article{nvidia2025nvfp4,
  author  = {{NVIDIA}},
  title   = {{Pretraining Large Language Models with NVFP4}},
  journal = {arXiv preprint arXiv:2509.25149},
  year    = {2025}
}

@article{micikevicius2022fp8,
  author  = {Micikevicius, Paulius and Stosic, Dusan and Burgess, Neil and Cornea, Marius and Dubey, Pradeep and Grisenthwaite, Richard and Ha, Sangwon and Heinecke, Alexander and Judd, Patrick and Kamalu, John and Mellempudi, Naveen and Oberman, Stuart and Shoeybi, Mohammad and Siu, Michael and Wu, Hao},
  title   = {{FP8 Formats for Deep Learning}},
  journal = {arXiv preprint arXiv:2209.05433},
  year    = {2022}
}

@misc{transformerengine,
  author       = {{NVIDIA}},
  title        = {{Transformer Engine}},
  year         = {2026},
  howpublished = {\url{https://github.com/NVIDIA/TransformerEngine}}
}

@misc{torchao,
  author       = {{PyTorch Team}},
  title        = {{TorchAO: PyTorch Architecture Optimization}},
  year         = {2026},
  howpublished = {\url{https://github.com/pytorch/ao}}
}

@inproceedings{dao2022flashattention,
  author    = {Dao, Tri and Fu, Daniel Y. and Ermon, Stefano and Rudra, Atri and R{\'e}, Christopher},
  title     = {{FlashAttention: Fast and Memory-Efficient Exact Attention with IO-Awareness}},
  booktitle = {Advances in Neural Information Processing Systems},
  volume    = {35},
  year      = {2022}
}

@inproceedings{zadouri2026fa4,
  author    = {Zadouri, Ted and Hoehnerbach, Markus and Shah, Jay and Liu, Timmy and Thakkar, Vijay and Dao, Tri},
  title     = {{FlashAttention-4: Algorithm and Kernel Pipelining Co-Design for Asymmetric Hardware Scaling}},
  booktitle = {Proceedings of Machine Learning and Systems},
  year      = {2026}
}

@article{dong2024flexattention,
  author  = {Dong, Juechu and Feng, Boyuan and Guessous, Driss and Liang, Yanbo and He, Horace},
  title   = {{Flex Attention: A Programming Model for Generating Optimized Attention Kernels}},
  journal = {arXiv preprint arXiv:2412.05496},
  year    = {2024}
}

@inproceedings{dettmers2022optimizers,
  author    = {Dettmers, Tim and Lewis, Mike and Shleifer, Sam and Zettlemoyer, Luke},
  title     = {{8-bit Optimizers via Block-wise Quantization}},
  booktitle = {International Conference on Learning Representations},
  year      = {2022}
}

@inproceedings{gupta2015limited,
  author    = {Gupta, Suyog and Agrawal, Ankur and Gopalakrishnan, Kailash and Narayanan, Pritish},
  title     = {{Deep Learning with Limited Numerical Precision}},
  booktitle = {Proceedings of the 32nd International Conference on Machine Learning},
  year      = {2015}
}

@inproceedings{ansel2024pytorch2,
  author    = {Ansel, Jason and others},
  title     = {{PyTorch 2: Faster Machine Learning Through Dynamic Python Bytecode Transformation and Graph Compilation}},
  booktitle = {Proceedings of the 29th ACM International Conference on Architectural Support for Programming Languages and Operating Systems},
  year      = {2024}
}

@article{chen2016checkpointing,
  author  = {Chen, Tianqi and Xu, Bing and Zhang, Chiyuan and Guestrin, Carlos},
  title   = {{Training Deep Nets with Sublinear Memory Cost}},
  journal = {arXiv preprint arXiv:1604.06174},
  year    = {2016}
}

@misc{mlx2023,
  author       = {Hannun, Awni and Digani, Jagrit and Katharopoulos, Angelos and Collobert, Ronan},
  title        = {{MLX: Efficient and Flexible Machine Learning on Apple Silicon}},
  year         = {2023},
  howpublished = {\url{https://github.com/ml-explore/mlx}}
}

@misc{turner2024mfa,
  author       = {Turner, Philip},
  title        = {{Metal FlashAttention}},
  year         = {2024},
  howpublished = {\url{https://github.com/philipturner/metal-flash-attention}}
}

@misc{garg2026imajev,
  author       = {Garg, Mohit},
  title        = {{imajev 1.0: Typed Decisions from Photos and App State}},
  year         = {2026},
  month        = sep,
  howpublished = {\url{https://mohit67890.github.io/imajev/report/}},
  note         = {Technical report}
}

@article{li2026edgejev,
  author  = {Li, Delong and Wang, Xu and Gong, Haochen and Lang, Rui and Yu, Guangsheng},
  title   = {{Replacing Large Language Models with Jev Decision Models for Low-Latency Edge Service Orchestration}},
  journal = {arXiv preprint arXiv:2609.22753},
  year    = {2026}
}

@misc{decisionbench2026,
  author       = {{Hanno Labs}},
  title        = {{DecisionBench 1.0}},
  year         = {2026},
  howpublished = {\url{https://huggingface.co/datasets/Hanno-Labs/decision-bench}}
}

@misc{imajevbench2026,
  author       = {Garg, Mohit},
  title        = {{ImajevBench v2.0-lite}},
  year         = {2026},
  howpublished = {\url{https://huggingface.co/datasets/mohit67890/imajev-bench}}
}

@inproceedings{kempka2016vizdoom,
  author    = {Kempka, Micha{\l} and Wydmuch, Marek and Runc, Grzegorz and Toczek, Jakub and Ja{\'s}kowski, Wojciech},
  title     = {{ViZDoom}: A {Doom}-based {AI} Research Platform for Visual Reinforcement Learning},
  booktitle = {IEEE Conference on Computational Intelligence and Games},
  pages     = {341--348},
  year      = {2016}
}

@misc{cirulli2014twenty48,
  author       = {Cirulli, Gabriele},
  title        = {2048},
  year         = {2014},
  howpublished = {\url{https://github.com/gabrielecirulli/2048}}
}

@misc{laya2026code,
  author       = {{Convai Innovations}},
  title        = {{Laya: Code and Benchmark Results}},
  year         = {2026},
  howpublished = {\url{https://github.com/NandhaKishorM/laya}}
}

@misc{jeff2026,
  author       = {Strasser, Mathias},
  title        = {{Jeff}},
  year         = {2026},
  howpublished = {\url{https://github.com/firelex/jeff}}
}

@misc{bakhta2026jevbenchmarks,
  author       = {Bakhta, Abdelhamid},
  title        = {{jev-benchmarks: Probability-Aware Evaluation for Typed Decision Models}},
  year         = {2026},
  howpublished = {\url{https://github.com/AbdelStark/jev-benchmarks}}
}

@misc{nibzard2026dmb,
  author       = {{nibzard}},
  title        = {{DMB: Decision-Model Benchmark}},
  year         = {2026},
  howpublished = {\url{https://github.com/nibzard/decision-model-benchmark}}
}

@inproceedings{casanueva2020banking77,
  author    = {Casanueva, I{\~n}igo and Tem{\v{c}}inas, Tadas and Gerz, Daniela and Henderson, Matthew and Vuli{\'c}, Ivan},
  title     = {{Efficient Intent Detection with Dual Sentence Encoders}},
  booktitle = {Proceedings of the 2nd Workshop on Natural Language Processing for Conversational AI},
  pages     = {38--45},
  year      = {2020}
}

@inproceedings{zhang2019paws,
  author    = {Zhang, Yuan and Baldridge, Jason and He, Luheng},
  title     = {{PAWS: Paraphrase Adversaries from Word Scrambling}},
  booktitle = {Proceedings of the 2019 Conference of the North American Chapter of the Association for Computational Linguistics: Human Language Technologies},
  pages     = {1298--1308},
  year      = {2019}
}

@inproceedings{socher2013sst,
  author    = {Socher, Richard and Perelygin, Alex and Wu, Jean and Chuang, Jason and Manning, Christopher D. and Ng, Andrew and Potts, Christopher},
  title     = {{Recursive Deep Models for Semantic Compositionality Over a Sentiment Treebank}},
  booktitle = {Proceedings of the 2013 Conference on Empirical Methods in Natural Language Processing},
  pages     = {1631--1642},
  year      = {2013}
}

@inproceedings{conneau2018xnli,
  author    = {Conneau, Alexis and Rinott, Ruty and Lample, Guillaume and Williams, Adina and Bowman, Samuel and Schwenk, Holger and Stoyanov, Veselin},
  title     = {{XNLI: Evaluating Cross-lingual Sentence Representations}},
  booktitle = {Proceedings of the 2018 Conference on Empirical Methods in Natural Language Processing},
  pages     = {2475--2485},
  year      = {2018}
}

@inproceedings{wang2019glue,
  author    = {Wang, Alex and Singh, Amanpreet and Michael, Julian and Hill, Felix and Levy, Omer and Bowman, Samuel R.},
  title     = {{GLUE: A Multi-Task Benchmark and Analysis Platform for Natural Language Understanding}},
  booktitle = {International Conference on Learning Representations},
  year      = {2019}
}

@inproceedings{fitzgerald2023massive,
  author    = {FitzGerald, Jack and Hench, Christopher and Peris, Charith and Mackie, Scott and Rottmann, Kay and Sanchez, Ana and Nash, Aaron and Urbach, Liam and Kakarala, Vishesh and Singh, Richa and Ranganath, Swetha and Crist, Laurie and Britan, Misha and Leeuwis, Wouter and Tur, Gokhan and Natarajan, Prem},
  title     = {{MASSIVE: A 1M-Example Multilingual Natural Language Understanding Dataset with 51 Typologically-Diverse Languages}},
  booktitle = {Proceedings of the 61st Annual Meeting of the Association for Computational Linguistics},
  pages     = {4277--4302},
  year      = {2023}
}

@inproceedings{saravia2018carer,
  author    = {Saravia, Elvis and Liu, Hsien-Chi Toby and Huang, Yen-Hao and Wu, Junlin and Chen, Yi-Shin},
  title     = {{CARER: Contextualized Affect Representations for Emotion Recognition}},
  booktitle = {Proceedings of the 2018 Conference on Empirical Methods in Natural Language Processing},
  pages     = {3687--3697},
  year      = {2018}
}

@inproceedings{clark2019boolq,
  author    = {Clark, Christopher and Lee, Kenton and Chang, Ming-Wei and Kwiatkowski, Tom and Collins, Michael and Toutanova, Kristina},
  title     = {{BoolQ: Exploring the Surprising Difficulty of Natural Yes/No Questions}},
  booktitle = {Proceedings of the 2019 Conference of the North American Chapter of the Association for Computational Linguistics: Human Language Technologies},
  pages     = {2924--2936},
  year      = {2019}
}

@inproceedings{barbieri2020tweeteval,
  author    = {Barbieri, Francesco and Camacho-Collados, Jose and Espinosa Anke, Luis and Neves, Leonardo},
  title     = {{TweetEval: Unified Benchmark and Comparative Evaluation for Tweet Classification}},
  booktitle = {Findings of the Association for Computational Linguistics: EMNLP 2020},
  pages     = {1644--1650},
  year      = {2020}
}

@inproceedings{welbl2017sciq,
  author    = {Welbl, Johannes and Liu, Nelson F. and Gardner, Matt},
  title     = {{Crowdsourcing Multiple Choice Science Questions}},
  booktitle = {Proceedings of the 3rd Workshop on Noisy User-generated Text},
  pages     = {94--106},
  year      = {2017}
}

@inproceedings{sakaguchi2020winogrande,
  author    = {Sakaguchi, Keisuke and Le Bras, Ronan and Bhagavatula, Chandra and Choi, Yejin},
  title     = {{WinoGrande: An Adversarial Winograd Schema Challenge at Scale}},
  booktitle = {Proceedings of the AAAI Conference on Artificial Intelligence},
  volume    = {34},
  pages     = {8732--8740},
  year      = {2020}
}

@inproceedings{sap2019socialiqa,
  author    = {Sap, Maarten and Rashkin, Hannah and Chen, Derek and Le Bras, Ronan and Choi, Yejin},
  title     = {{Social IQa: Commonsense Reasoning about Social Interactions}},
  booktitle = {Proceedings of the 2019 Conference on Empirical Methods in Natural Language Processing and the 9th International Joint Conference on Natural Language Processing},
  pages     = {4463--4473},
  year      = {2019}
}

@inproceedings{zhang2015agnews,
  author    = {Zhang, Xiang and Zhao, Junbo and LeCun, Yann},
  title     = {{Character-level Convolutional Networks for Text Classification}},
  booktitle = {Advances in Neural Information Processing Systems},
  volume    = {28},
  year      = {2015}
}

@inproceedings{niu2024ragtruth,
  author    = {Niu, Cheng and Wu, Yuanhao and Zhu, Juno and Xu, Siliang and Shum, KaShun and Zhong, Randy and Song, Juntong and Zhang, Tong},
  title     = {{RAGTruth: A Hallucination Corpus for Developing Trustworthy Retrieval-Augmented Language Models}},
  booktitle = {Proceedings of the 62nd Annual Meeting of the Association for Computational Linguistics},
  pages     = {10862--10878},
  year      = {2024}
}

@inproceedings{tan2025judgebench,
  author    = {Tan, Sijun and Zhuang, Siyuan and Montgomery, Kyle and Tang, William Y. and Cuadron, Alejandro and Wang, Chenguang and Popa, Raluca Ada and Stoica, Ion},
  title     = {{JudgeBench: A Benchmark for Evaluating LLM-Based Judges}},
  booktitle = {International Conference on Learning Representations},
  year      = {2025}
}

@inproceedings{lin2023toxicchat,
  author    = {Lin, Zi and Wang, Zihan and Tong, Yongqi and Wang, Yangkun and Guo, Yuxin and Wang, Yujia and Shang, Jingbo},
  title     = {{ToxicChat: Unveiling Hidden Challenges of Toxicity Detection in Real-World User-AI Conversation}},
  booktitle = {Findings of the Association for Computational Linguistics: EMNLP 2023},
  pages     = {4694--4702},
  year      = {2023}
}

@inproceedings{nguyen2016msmarco,
  author    = {Nguyen, Tri and Rosenberg, Mir and Song, Xia and Gao, Jianfeng and Tiwary, Saurabh and Majumder, Rangan and Deng, Li},
  title     = {{MS MARCO: A Human Generated MAchine Reading COmprehension Dataset}},
  booktitle = {Proceedings of the Workshop on Cognitive Computation: Integrating Neural and Symbolic Approaches},
  year      = {2016}
}

@inproceedings{metsis2006spam,
  author    = {Metsis, Vangelis and Androutsopoulos, Ion and Paliouras, Georgios},
  title     = {{Spam Filtering with Naive Bayes -- Which Naive Bayes?}},
  booktitle = {Third Conference on Email and Anti-Spam},
  year      = {2006}
}

@article{cobbe2021gsm8k,
  author  = {Cobbe, Karl and Kosaraju, Vineet and Bavarian, Mohammad and Chen, Mark and Jun, Heewoo and Kaiser, Lukasz and Plappert, Matthias and Tworek, Jerry and Hilton, Jacob and Nakano, Reiichiro and Hesse, Christopher and Schulman, John},
  title   = {{Training Verifiers to Solve Math Word Problems}},
  journal = {arXiv preprint arXiv:2110.14168},
  year    = {2021}
}

@article{austin2021mbpp,
  author  = {Austin, Jacob and Odena, Augustus and Nye, Maxwell and Bosma, Maarten and Michalewski, Henryk and Dohan, David and Jiang, Ellen and Cai, Carrie and Terry, Michael and Le, Quoc and Sutton, Charles},
  title   = {{Program Synthesis with Large Language Models}},
  journal = {arXiv preprint arXiv:2108.07732},
  year    = {2021}
}

@misc{deepset2023injections,
  author       = {{deepset}},
  title        = {{prompt-injections}},
  year         = {2023},
  howpublished = {\url{https://huggingface.co/datasets/deepset/prompt-injections}}
}

@misc{liu2024phishing,
  author       = {Liu, Zefang},
  title        = {{Phishing Email Dataset}},
  year         = {2024},
  howpublished = {\url{https://huggingface.co/datasets/zefang-liu/phishing-email-dataset}}
}

@misc{bueck2025tickets,
  author       = {{Tobi-Bueck}},
  title        = {{Customer Support Tickets}},
  year         = {2025},
  howpublished = {\url{https://huggingface.co/datasets/Tobi-Bueck/customer-support-tickets}}
}

\clearpage
\appendix
\crefalias{section}{appendix}
\section{Interface details}\label{app:interface}

Each question compiles to a separate decoder sequence. The sequence for a Choice question is:
\begin{lstlisting}
<bos>type: choice
instructions: "Which stock category applies?"
candidate: {"name":"empty","description":"No items."}<candidate_end>
candidate: {"name":"low","description":"1-7 items."}<candidate_end>
candidate: {"name":"high","description":"8+ items."}<candidate_end>
<decision_end>
\end{lstlisting}
Noul uses the candidates \texttt{true} and \texttt{false}, and Score lists its levels in order. Choice confidence is the margin of the top probability over the uniform probability, normalised to $[0, 1]$: $(p_{\max} - 1/K)/(1 - 1/K)$. Score confidence is $1 - d/d_u$, where $d$ is the mean distance of the distribution from its mode and $d_u$ is the mean distance of a uniform distribution from the same mode. The following listing shows a request and an illustrative response.
\begin{lstlisting}
{"state": {"available": 7},
 "questions": {
   "enough": {"type": "noul", "instructions": "Are at least 8 items available?"},
   "stock":  {"type": "choice", "instructions": "Which stock category applies?",
              "criteria": {"empty": "No items are available.",
                           "low":   "Between 1 and 7 items are available.",
                           "high":  "At least 8 items are available."}},
   "level":  {"type": "score", "instructions": "Rate the available stock on this scale.",
              "criteria": ["No items are available.",
                           "Between 1 and 7 items are available.",
                           "At least 8 items are available."]}}}

{"model": "bongard-mini",
 "answers": {
   "enough": {"type": "noul", "noul": 0.03},
   "stock":  {"type": "choice", "choice": "low", "confidence": 0.94,
              "probabilities": {"empty": 0.01, "low": 0.96, "high": 0.03}},
   "level":  {"type": "score", "score": 1.03, "confidence": 0.925,
              "probabilities": {"0": 0.01, "1": 0.95, "2": 0.04},
              "legend": {"0": "No items are available.",
                         "1": "Between 1 and 7 items are available.",
                         "2": "At least 8 items are available."}}},
 "usage": {"input_tokens": 212, "output_tokens": 0}}
\end{lstlisting}

\section{Stage-1 task families}\label{app:data}

Stage~1 covers language and document judgments, structured reasoning, retrieval, images, computer use, operational decisions and probability questions. Depending on the task, targets come from annotations, executable rules, exact mechanisms or filtered teacher judgments. Related examples remain in the same split.

\section{Stage-2 details}\label{app:jepa}

\noindent\begin{minipage}{\linewidth}
  \centering
  \captionsetup{type=table}\caption{Relation families. Separately encoded families carry the content objective, and the answer-state family carries the answer contrast. Other shared-state families receive supervision on both views but no alignment term.}
  \label{tab:relations}
  \small
  \begin{tabularx}{\linewidth}{@{}P{3.3cm}Lll@{}}
    \toprule
    Family & View A $\leftrightarrow$ view B & Encoding & Alignment \\
    \midrule
    Query $\to$ passage & Query with a relevance task $\leftrightarrow$ the passage's content & Separate & Content \\
    Question $\to$ answer & Question over evidence $\leftrightarrow$ the answer's content & Separate & Content \\
    Image $\to$ content & Image with a scoped prediction $\leftrightarrow$ its annotated content & Separate & Content \\
    World prediction & World and action $\leftrightarrow$ observed next state & Separate & Content \\
    Causal prediction & Causal model and intervention $\leftrightarrow$ observed outcome & Separate & Content \\
    Assignment and matching & Problem instance $\leftrightarrow$ verified allocation & Separate & Content \\
    Paraphrase & Sentence $\leftrightarrow$ meaning-preserving rewrite & Separate & Content \\
    Program $\to$ result & Program and input $\leftrightarrow$ re-executed result & Separate & Content \\
    Answer state & Multiple-choice question $\leftrightarrow$ correct option (yes) and wrong option (no) & Shared & Answer \\
    Logical complement & Noul question $\leftrightarrow$ its negation & Shared & None \\
    Inherited view & Verified equivalent views & Shared & None \\
    \bottomrule
  \end{tabularx}
\end{minipage}

\medskip
\textbf{Training set and schedule.} We balance relation families and replay earlier tasks to limit drift. Held-out split groups are selected before view construction, so related origins do not cross the split. The alignment weight warms up over the first 5\% of updates. A pack with fewer than four members of a kind skips the contrast for that kind.

\textbf{Panel details.} On the frozen panel, conditional-event accuracy for the correct option's ``yes'' rose from 64.2\% to 85.3\%. For a wrong option's ``no'', it rose from 85.4\% to 92.8\%. Image-evidence invariance improved: the total variation fell from 0.017 to 0.006.

\section{Sandbox environments}\label{app:envs}

Stage~3 uses games, grid worlds, verifiable judgments, business workflows, computer use, classification and executable generators. States can be text, structured records or accessibility snapshots. The model answers typed questions about candidate actions. Outcome labels come from exact computation, recorded observations or rollouts under the stated continuation policy.

\section{Training-stack details}\label{app:systems}

\textbf{NVFP4 feed-forward blocks.} NVFP4 is a 4-bit floating-point format with two-level block scaling, and Blackwell GPUs execute it natively \citep{nvidia2025nvfp4}. The model has 68 feed-forward blocks, 34 per stack. Each block is a fused Transformer Engine module that reuses the original parameters and maps back to the standard checkpoint keys on save. In BF16, the fused block reproduces the original module to a relative $L_2$ error of 0.0036. Saved checkpoints load in the standard Transformers implementation with identical logits. In stage~3, the feed-forward projections run as NVFP4 Transformer Engine linear layers instead, with one matrix multiplication for the gate and up projections. Its checkpoints also use the standard layout.

\textbf{FP8 attention projections.} In stages 1 and 2, the 272 attention projections, four per layer, are FP8 matrix multiplications with rowwise scaling. Attention arithmetic stays in BF16, because the FlashAttention-4 kernels for head dimension 256 in variable-length layouts do not support FP8 training.

\textbf{Packing.} Records are packed without padding into streams of states and questions, and positions restart at zero for each sequence. In the encoder, each state attends only to its own tokens. In the decoder, each question attends only to its own prefix and its own state. Sequences long enough to trigger the 1,024-token sliding window go through FlexAttention \citep{dong2024flexattention}. Long and short sequences are split only inside attention.

\textbf{Memory.} Parameters, gradients, embeddings and residuals are stored in BF16. The 8-bit AdamW state takes one byte per parameter per moment. Stochastic rounding keeps small updates from vanishing without an FP32 master copy. State keys and values are stored once per record and gathered again for each question in the backward pass, with shared gradients accumulated in FP32. Only records longer than the pack limit use per-layer recomputation, so no input is truncated. Peak allocation was 179~GB in stage~2 and 239~GB in the warm start of stage~3.

\textbf{Kernel overhead.} We batch the readout indexing and the finiteness checks per update. This change cut the kernel count of a steady-state pack from 21,528 to 17,948. Tokeniser settings, compiled requests and readout positions are stored with the data, so training reads precomputed tokens and token costs.

\textbf{Other hardware.} The CPU and Apple Silicon paths share the model code. For smaller machines, a LoRA configuration applies rank-8 adapters to the attention projections and also trains the projector and the head.

\section{Same-item comparison}\label{app:items}

Laya and Jeff item sets are rebuilt from their published code, with their states, instructions and candidates \citep{laya2026code,jeff2026}. Elsewhere, one sentence states the task, and the candidates are the benchmark's own labels. Yes/no tasks are Noul questions (\cref{tab:items}).

\begin{table}[!ht]
  \centering
  \caption{Item sets and accuracy of the same-item comparison, in the order of \cref{fig:same-items}. Test and validation splits are complete.}
  \label{tab:items}
  \footnotesize\renewcommand{\arraystretch}{0.94}
  \begin{tabularx}{\linewidth}{@{}Llrrr@{}}
    \toprule
    & & & \multicolumn{2}{c@{}}{Accuracy (\%)} \\
    \cmidrule(l){4-5}
    Benchmark & Item set & Items & Bongard & Jev \\
    \midrule
    Banking77 \citep{casanueva2020banking77} & test & 3,076 & 93.1 & 79.8 \\
    Support-ticket triage \citep{bueck2025tickets} & Laya & 400 & 43.0 & 35.8 \\
    PAWS \citep{zhang2019paws} & test & 8,000 & 91.1 & 84.9 \\
    SST-5 \citep{socher2013sst} & Laya & 600 & 58.2 & 57.5 \\
    Model routing \citep{cobbe2021gsm8k,austin2021mbpp,zhang2015agnews} & Laya & 399 & 98.0 & 97.7 \\
    XNLI, 15 languages \citep{conneau2018xnli} & Laya & 4,500 & 74.6 & 74.3 \\
    Long documents \citep{laya2026code} & Laya & 260 & 100.0 & 100.0 \\
    RAG passage relevance \citep{nguyen2016msmarco} & Laya & 400 & 60.2 & 61.3 \\
    QNLI \citep{wang2019glue} & validation & 5,463 & 92.2 & 93.7 \\
    MASSIVE scenario, 14 languages \citep{fitzgerald2023massive} & Laya & 4,200 & 69.0 & 70.6 \\
    Emotion \citep{saravia2018carer} & test & 2,000 & 57.7 & 59.3 \\
    BoolQ \citep{clark2019boolq} & Laya & 600 & 89.2 & 91.5 \\
    TweetEval offensive \citep{barbieri2020tweeteval} & test & 860 & 80.7 & 83.1 \\
    Jailbreak detection \citep{lin2023toxicchat} & Laya & 400 & 92.2 & 94.8 \\
    Phishing email \citep{liu2024phishing} & Laya & 400 & 87.8 & 90.2 \\
    SciQ \citep{welbl2017sciq} & test & 1,000 & 95.5 & 99.1 \\
    Email spam \citep{metsis2006spam} & Laya & 400 & 91.8 & 97.2 \\
    WinoGrande \citep{sakaguchi2020winogrande} & validation & 1,267 & 85.7 & 91.3 \\
    AG News \citep{zhang2015agnews} & test & 7,600 & 82.2 & 88.4 \\
    RAGTruth \citep{niu2024ragtruth} & Jeff & 1,500 & 74.9 & 82.8 \\
    MASSIVE intent, 51 languages \citep{fitzgerald2023massive} & Laya & 5,100 & 77.4 & 89.3 \\
    Toxicity \citep{lin2023toxicchat} & Laya & 400 & 53.5 & 67.5 \\
    Prompt injection \citep{deepset2023injections} & Laya & 116 & 56.9 & 75.9 \\
    JudgeBench \citep{tan2025judgebench} & Jeff & 350 & 55.4 & 79.7 \\
    \bottomrule
  \end{tabularx}
\end{table}

\FloatBarrier

\section{Contributions and acknowledgements}\label{app:contrib}

\textbf{Authors.} Li Ding, Haidi Jin and Chen Ji (AgentBull Pte Ltd).

\textbf{Contributions.} Li Ding conceived the project, designed the architecture and the training programme, wrote the code, ran all training and evaluation, and wrote the report. Haidi Jin and Chen Ji processed the training datasets.

\textbf{Acknowledgements.} We thank Google for the open T5Gemma~2 weights, and the maintainers of JevBench, DecisionBench, typed-decisions, behavior-benchmark and ImajevBench for their public benchmarks and results. We also thank the authors of Laya, Jeff, jev-benchmarks and DMB for their open benchmark code. Li Ding is also grateful to Douglas Richard Hofstadter, whose writings and ideas inspired his thinking and whose books introduced him to the Bongard problems that gave the model its name. He thanks his wife, Angela Chan, and his dog, Nieh-Nieh, for their companionship throughout this work.

\end{document}